\documentclass{article}
\usepackage{arxiv}

\usepackage[utf8]{inputenc} 
\usepackage[T1]{fontenc}    
\usepackage{hyperref}       
\usepackage{url}            
\usepackage{booktabs}       
\usepackage{amsfonts}       
\usepackage{amsmath}
\usepackage{nicefrac}       
\usepackage{microtype}      
\usepackage{lipsum}
\usepackage{graphicx}
\usepackage{float}
\usepackage{caption}
\usepackage{titlesec}
\usepackage{natbib}
\usepackage{setspace}
\usepackage{algorithm}
\usepackage{algpseudocode}
\usepackage{booktabs}
\usepackage{makecell}
\titlespacing*{\section}{8pt}{12pt}{8pt}
\titlespacing*{\subsection}{8pt}{8pt}{8pt}
\titlespacing*{\subsubsection}{8pt}{8pt}{8pt}

\graphicspath{ {./images/} }
\usepackage{authblk}

\usepackage{fancyvrb}

\DefineVerbatimEnvironment{CodeInput}{Verbatim}{fontshape=sl}
\DefineVerbatimEnvironment{CodeOutput}{Verbatim}{}
\newcommand{\pkg}[1]{{\fontseries{m}\fontseries{b}\selectfont #1}}
\let\proglang=\textsf

\title{PyDTS: A Python Package for Discrete-Time Survival Analysis with Competing Risks and Optional Penalization}

\author[1]{Tomer Meir$^{*}$}
\author[1]{Rom Gutman$^{*}$}
\author[2]{Malka Gorfine}

\affil[1]{Faculty of Data and Decision Sciences, Technion - Israel Institute of Technology}
\affil[2]{Department of Statistics and Operations Research, Tel Aviv University}
\affil[*]{Corresponding Authors: tomer1812@gmail.com, rom.gutman1@gmail.com}
\begin{document}

\maketitle



\begin{abstract}
Time-to-event (survival) analysis models the time until a pre-specified event occurs. When time is measured in discrete units or rounded into intervals, standard continuous-time models can yield biased estimators. In addition, the event of interest may belong to one of several mutually exclusive types, referred to as competing risks, where the occurrence of one event prevents the occurrence or observation of the others. \pkg{PyDTS} is an open-source Python package for analyzing discrete-time survival data with competing-risks. It provides regularized estimation methods, model evaluation metrics, variable screening tools, and a simulation module to support research and development.
\end{abstract}


\section{Introduction}
\label{sec:intro}
Discrete-time survival data pertains to situations where data are limited to a discrete grid, either because events only happen at regular discrete time points, or because the available data only record the interval during which each event occurred. For example,  cancer deaths may be measured in months since diagnosis \citep{lee_analysis_2018}, and hospital stays are recorded in days. Most regression models for survival data assume that time is continuous. However, applying standard continuous-time models to discrete-time data may lead to substantially biased estimators for the discrete-time models \citep{lee_analysis_2018, wu2022analysis}.

Here we consider the setting of discrete-time survival data with competing events in which subjects can experience only one of several different types of events over follow-up. For example, competing risks of hospital length of stay, measured in days, are discharge and in-hospital death. Occurrence of one of these events precludes us from observing the other event of the same patient. Cause-specific mortality is another typical example of competing risks, where individuals could pass away due to various causes, such as heart disease, cancer, or other reasons \citep{kalbfleisch_statistical_2011,klein_survival_2003}. 

When dealing with continuous-time survival data with competing risks, the classical methods for analyzing non-competing events can be employed, since the likelihood function for the continuous-time setting can be factored into distinct likelihoods for each cause-specific hazard function \citep{kalbfleisch_statistical_2011}. However, this approach does not hold true for discrete-time data with competing risks, as highlighted in \cite{lee_analysis_2018} and references therein.  Most existing works on discrete-time data with competing risks are based on simultaneously estimating all the parameters via the full likelihood function, which are computationally time consuming. On the other hand,
 recent works \citep{lee_analysis_2018, schmid2021competing} have shown that a collapsed-likelihood approach may be used to separate estimation of cause-specific hazard models for different event types. This approach is equivalent to fitting a generalized linear model (GLM) to repeated binary outcomes, and yields consistent and asymptotically normal estimators under standard regularity conditions. In addition, Wald-type confidence intervals and likelihood-ratio tests may be used to evaluate the effects of covariates.

Recently, \cite{meir_gorfine_dtsp_2023} developed a new estimation procedure  for a semi-parametric logit-link survival model for discrete time with competing events and right censoring. Their procedure has two main advantages: (i) It is significantly faster than existing methods and requires a smaller amount of memory resources. For example, with 20,000 observations, 10 covariates, and two competing events, the computation time reduction is a factor of 5 compared to that of \cite{lee_analysis_2018}.
(ii) It allows including modern machine-learning model-selection procedures, such as regularization and screening.

To the best of our knowledge, there is only one \proglang{R} package for discrete-time survival data with competing events named \pkg{discSurv} \citep{discSurv22}. Two main functions of this package  are relevant in the context of competing events: (i) \texttt{compRisksGEE} applies the estimation procedure of \cite{lee_analysis_2018} with the logit-link function. (ii) \texttt{survTreeLaplaceHazard} predicts the Laplace-smoothed hazards of a discrete-survival tree without or with competing events. This function uses different cause-specific hazard functions; for details see \cite{berger2019classification}.

The aim of this work is to present \pkg{PyDTS}, which is currently the sole \proglang{Python} software package that deals with discrete-time survival data with competing events. Additionally, it is the first package to incorporate a regularized regression approach for such data. The features of \pkg{PyDTS} comprise of:
\begin{itemize}
    \item Implementation of the estimation techniques of \cite{lee_analysis_2018}  and \cite{meir_gorfine_dtsp_2023} under the logit-link function.
    \item Computation of performance metrics, such as the area under the  receiver operating characteristic  curve and Brier score, for discrete-time survival data with competing events.
    \item Implementation of regularized regression based on \cite{meir_gorfine_dtsp_2023} with K-fold CV.
    \item Generation of simulated discrete-time survival data with competing events and right censoring.
\end{itemize}

The rest of the paper is organized as follows. Section \ref{sec:methods} describes the model and estimation techniques applied in \pkg{PyDTS}. Section \ref{sec:datageneration} demonstrates how to simulate discrete-time survival data with competing events based on \pkg{PyDTS}. Section \ref{ref:illustations} provides various examples of \pkg{PyDTS} usage. Section \ref{sec:mimic-los} demonstrates the package utility by analysing length of hospital stay based on the Medical Information Mart for Intensive Care (MIMIC) - IV dataset. Some concluding remarks are provided in Section \ref{sec:summary} .

\section{Model and Methods}
\label{sec:methods}

\subsection{Notation and Model}
Assume $T$ is a discrete random variable of event time that can take on only the values $\{1,2,\ldots,d\}$ and $J$ denotes the type of event, $J \in \{1,\ldots,M\}$.  Consider a $p \times 1$ vector of time-independent covariates $Z$. The setting of time-dependent covariates will be discussed later. A general discrete cause-specific hazard function is of the form
$$
\lambda_j(t|Z) = \Pr(T=t,J=j|T\geq t, Z)  \hspace{0.1cm} , \hspace{0.3cm} t =1,2,\ldots,d  \hspace{0.1cm} , \hspace{0.3cm} j=1,\ldots,M  \, .
$$
Following \cite{allison_discrete-time_1982}, the logit-link semi-parametric models of the above hazard functions are given by
\begin{equation}\label{eq:logis}
\lambda_j(t|Z)=\frac{\exp(\alpha_{jt}+Z^T\beta_j)}{1+\exp(\alpha_{jt}+Z^T\beta_j)} \hspace{0.1cm} , \hspace{0.3cm} t =1,2,\ldots,d  \hspace{0.1cm} , \hspace{0.3cm} j=1,\ldots,M  \, ,
\end{equation}
where 
$$ 
\Omega = (\alpha_{11},\ldots,\alpha_{1d},\beta_1^T,  \ldots,  \alpha_{M1},\ldots,\alpha_{Md},\beta_M^T)
$$
is an unknown vector of parameters. The total number of unknown parameters is $M(d+p)$.  Leaving $\alpha_{jt}$ unspecified is analogous to an unspecified baseline hazard function in the Cox proportional hazard model \citep{cox_regression_1972}, and thus the above model is considered as a semi-parametric discrete-time model.
For simplicity of presentation, the vector of covariates $Z$ is shared by the $M$ models. However, it does not imply that identical covariates must be shared by the models, since the regression coefficient vectors $\beta_j$ are event-type dependent, and any coefficient of $\beta_j$ can be set to 0 in order to exclude the corresponding covariate.  One of our goals is estimating  
$\Omega$.

Assume the data consist of $n$ independent observations, each with $(X_i,\delta_i,J_i,Z_i)$ where $X_i=\min(C_i,T_i)$, $C_i$ is a right-censoring time, 
$\delta_i=I(T_i \leq C_i)$ is the event indicator and $J_i\in\{0,1,\ldots,M\}$, where $J_i=0$ if and only if $\delta_i=0$,  $i=1,\ldots,n$. It is assumed that given the covariates, the censoring and failure times are independent and non-informative. 

\subsection{The Collapsed Log-Likelihood Approach of Lee et al.}
\label{subsection:lee2018}
\cite{lee_analysis_2018} showed that in contrast to the continuous-time setting with competing events, the likelihood function of the discrete-time setting cannot be decomposed into separate likelihoods for each cause-specific
hazard function $\lambda_j$. This implies that estimating 
$\Omega$ by maximizing the likelihood, would require maximization with respect to $M(d+p)$ parameters simultaneously, which is expected to be time consuming. 

Alternatively, \cite{lee_analysis_2018}  suggested the following collapsed log-likelihood. 
The dataset is expanded such that for each observation $i$ the expanded dataset includes $X_i$ rows, i.e., pseudo observations, one row for each time $t$, $t \leq X_i$; see Table~\ref{tbl:expanded} for $M=2$ competing events. At each time point $t$ the pseudo observations can be viewed as random variables from a conditional multinomial distribution with one of $M+1$ possible outcomes $\{\delta_{1it},\ldots,\delta_{Mit},1-\sum_{j=1}^{M}\delta_{jit}\}$, where $\delta_{jit}$ equals 1 if individual $i$ experienced event of type $j$ at time $t$; and 0 otherwise. Then, for $M$ competing events,  the estimators of $(\alpha_{j1},\ldots,\alpha_{jd},\beta_j^T)$, $j=1,\ldots,M$, are the  values that maximize  
\begin{equation}\label{eq:logLj}
\log L_j = \sum_{i=1}^n \sum_{t=1}^{X_i}\left[ \delta_{jit} \log \lambda_j(t|Z_i)+(1-\delta_{jit})\log \{1-\lambda_j(t|Z_i)\} \right] \quad j=1,\ldots,M \, .
\end{equation}
Namely, each maximization $j$, $j=1,\ldots,M$, consists of  maximizing $d + p$ parameters simultaneously. Lee et al. showed that the estimators  are asymptotically  multivariate normally distributed and the covariance matrix can be consistently estimated.

\subsection{The Rapid Two-Step Approach}
\cite{meir_gorfine_dtsp_2023} adopted the collapsed log-likelihood approach but suggested 
 estimating each $\beta_j$, $j=1,\ldots,M$, separately from $\alpha_{jt}$. Their procedure improves over \cite{lee_analysis_2018} in two aspects: (1) By reducing the computation time substantially, especially for settings with large values of $d$. (2) It enables easy incorporation of penalization regression methods (e.g., ridge, lasso and elastic net among others) and screening methods. 

Let $\tilde{X}$ be the new column of times of the expanded dataset, as demonstrated in Table \ref{tbl:expanded}. For each event type $j$, $\beta_j$ is estimated by a conditional logistic regression model,  while stratifying the expanded dataset according to $\tilde{X}$ and conditioning on the number of events within each stratum. Hence, the estimators of $\beta_j$, $j=1,\ldots,M$, are obtained by maximizing
\begin{equation}\label{ref:betaj}
L_j^{\mathcal{C}}(\beta_j)  =  \prod_{t=1}^{d} \frac{\exp(\sum_{i \in \mathcal{C}_t} \delta_{jit} Z_i^T \beta_j)}{\sum_{d_{jt} \in \mathcal{S}_t} \exp(\sum_{i \in \mathcal{C}_t} d_{jit} Z_i^T \beta_j)}
\,\,\, , \,\,\, j=1,\ldots,M \, ,    
\end{equation}
where $\mathcal{C}_t$ is the set of all pseudo observations with $\tilde{X}$ equals $t$,  $\mathcal{S}_t$ is the set of all possible combinations of $\sum_{i=1}^n \delta_{jit}$ ones and $ \sum_{i=1}^n (1-\delta_{jit})$ zeros, $d_{jt}$ is a vector in $\mathcal{S}_t$, $d_{jit}$  equals to 0 or 1 with $\sum_{i}\delta_{jit}=\sum_{i}d_{jit}$, and $d_{jit}$ is a component of $d_{jt}$.
Since  Equation (\ref{ref:betaj}) has a form of partial likelihood of a Cox regression model when ties are present (see, for example,  Equation (8.4.3) of \cite{klein_survival_2003}), an available Cox model routine can be used for estimating $\beta_j$, $j=1,\ldots,M$. 

Given the estimators of $\beta_j$, $\widehat{\beta}_j$, $j=1,\ldots,M$,  $\alpha_{jt}$, $j=1,\ldots,M$, $t=1,\ldots,d$, are estimated by a series of $Md$ one-dimension simple optimization algorithms applied on the original dataset, such that 
    \begin{equation}\label{eq:alpha}
        \widehat{\alpha}_{jt} = 
        \mbox{argmin}_{a} \left\{ \frac{1}{Y.(t)} \sum_{i=1}^n I(X_i \geq t)\frac{\exp(a+Z_i^T\widehat{\beta}_j)}{1+\exp(a+Z_i^T\widehat{\beta}_j)} - \frac{N_j(t)}{Y.(t)}\right\}^2
    \end{equation}
where $Y.(t)=\sum_{i=1}^n I(X_i \geq t)$ and $N_j(t)=\sum_{i=1}^n I(X_i = t, J_i=j)$. The two-step estimation procedure of \cite{meir_gorfine_dtsp_2023} consists of the two speedy steps described in Algorithm 1. 

\begin{algorithm}
\caption{The Two-Step Estimation Procedure of \cite{meir_gorfine_dtsp_2023}}\label{alg:cap}
\begin{enumerate}
    \item Use the expanded dataset, estimate each vector $\beta_j$ separately, $j=1,\ldots, M$, by maximizing  Equation~(\ref{ref:betaj}) with a stratified Cox routine, and get $\widehat{\beta}_j$, $j=1,\ldots, M$.
    \item Given $\widehat{\beta}_j$ , $j=1,\ldots, M$, use the original non-expanded dataset and estimate each $\alpha_{jt}$, $j=1,\ldots,M$, $t=1,\ldots,d$, separately, by  Equation~(\ref{eq:alpha}).
\end{enumerate}
\end{algorithm}

The simulation results of \cite{meir_gorfine_dtsp_2023} reveal that the above two-step procedure performs well in terms of bias, and provides similar standard errors to that of \cite{lee_analysis_2018}. However, under large values of $d$,  a substantial improvement in computational time is achieved  by using the above two-step procedure. Apparently, estimating $p+d$ parameters simultaneously is more time consuming than estimating $d$ one-dimensional parameters and one parameter of $p$ dimension. 

\subsection{Time-dependent covariates}
Similarly to the continuous-time Cox model, the simplest way to
code time-dependent covariates uses intervals of time \citep{therneau2000cox}. Then, the data is encoded by breaking the individual’s
time into multiple time intervals, with one row of data for each interval. Hence combining this data expansion
step with the expansion demonstrated in Table \ref{tbl:expanded} is straightforward.

\subsection{Regularized Regression Models}
Penalized regression methods, such as LASSO, adaptive LASSO, elastic net~\citep{hastie2009elements}, place a constraint on the size of the regression coefficients. The estimation procedure of \cite{meir_gorfine_dtsp_2023} that separates the estimation  of $\beta_j$ and $\alpha_{jt}$ can easily incorporate such constraints  in Lagrangian form  by minimizing  
\begin{equation}\label{ref:betajregul}
-\log L_j^c(\beta_j)  + \eta_j P(\beta_j)
\,\,\, , \,\,\, j=1,\ldots,M \, ,    
\end{equation}
where $P$ is a penalty function and $\eta_j \geq 0$ are shrinkage tuning parameters. The parameters $\alpha_{jt}$ are estimated once the regularization step is completed and $\beta_j$ were estimated.   Clearly, any regularized Cox regression model routines can be used for estimating  $\beta_j$, $j=1,\ldots,M$, based on  Equation (\ref{ref:betajregul}), for example, the \texttt{CoxPHFitter} of \pkg{Lifelines} \citep{davidson-pilon_lifelines_2019} with penalization.

\subsection{Performance Measures}
\label{sec:performance}
Based on \cite{meir_gorfine_dtsp_2023}, \pkg{PyDTS} includes estimates of cause-specific incidence/dynamic area under the receiver operating characteristics curve (AUC) and Brier score (BS) for discrete survival data with competing events and right censoring. 

The cause-specific AUC at time $t$ is defined as the
probability of a random observation with observed event $j$ at time $t$ having a higher risk prediction for cause
$j$ than a randomly selected observation at risk at time $t$, free of  event $j$ at time $t$. Formally, 
\begin{equation*}
\mbox{AUC}_j(t) = \Pr (\pi_{ij}(t) > \pi_{mj}(t) \mid D_{ij} (t) = 1, D_{mj} (t) = 0, T_m \geq t) \,\,\, j=1,\ldots,M \,\,\, t=1,\ldots,d
\end{equation*}
where 
\begin{equation*}
\label{eq:piij}
\pi_{ij}(t) = \widehat{\Pr}(T_i=t, J_i=j \mid Z_i) = \widehat{\lambda}_j (t \mid Z_i) \widehat{S}(t-1 \mid Z_i) \, ,
\end{equation*}
$$
\widehat{\lambda}_j(t|Z)=\frac{\exp(\widehat{\alpha}_{jt}+Z^T \widehat{\beta}_j)}{1+\exp(\widehat{\alpha}_{jt}+Z^T \widehat{\beta}_j)} \, ,
$$
$$
\widehat{S}(t|Z) = \prod_{k=1}^{t} \left\{  1 - \sum_{j=1}^M \widehat{\lambda}_j(k|Z) \right\} \, ,
$$
and
\begin{equation*}
D_{ij} (t) = I(T_i = t, J_i = j) \, .
\end{equation*}
As explained in \cite{meir_gorfine_dtsp_2023}, the $\mbox{AUC}_j (t)$ can be estimated by
\begin{equation*}
\widehat{\mbox{AUC}}_j (t) 
=  \frac{\sum_{i=1}^{n}\sum_{m=1}^{n} D_{ij}(t)\bar{D}_{mj}(t)I(X_m \geq t) 
\{I(\pi_{ij}(t) > \pi_{mj}(t))+0.5I(\pi_{ij}(t)=\pi_{mj}(t))\}}{\sum_{i=1}^{n}\sum_{m=1}^{n} D_{ij}(t)\bar{D}_{mj}(t)I(X_m \geq t)}
\end{equation*}
$j=1,\ldots,M$, $t=1,\ldots,d$ where $\bar{D}_{ij}(t)=1-D_{ij}(t)$. An estimator of a cause-specific time-independent AUC is defined by
\begin{equation*}
\widehat{\mbox{AUC}}_j = \sum_{t=1}^d w_j (t) \widehat{\mbox{AUC}}_j (t) \,\,\, j=1,\ldots,M
\end{equation*}
where 
$$
w_j(t) = \frac{N_j(t)}{\sum_{t=1}^d N_j(t)} \, .
$$  
Finally, an estimator of the global AUC is defined as
\begin{equation*}
\widehat{\mbox{AUC}} = \sum_{j=1}^M v_j \widehat{\mbox{AUC}}_j 
\end{equation*}
where 
$$
v_j = \frac{\sum_{t=1}^d N_j(t)}{ \sum_{j=1}^M \sum_{t=1}^{d} N_j(t) } \, .
$$

Another well-known performance measurement is BS, a metric  measures the accuracy of probabilistic forecasts. The cause-specific BS at time $t$ is defined as
\begin{equation*}\label{ref:prederr}
  \widehat{\mbox{BS}}_{j}(t) = \frac{1}{Y_{\cdot}(t)} {\sum_{i=1}^n W_{ij}(t) \left\{ D_{ij}(t) - \pi_{ij}(t)\right\}^2}   \,\,\, j=1,\ldots,M \,\,\, t=1,\ldots,d
\end{equation*}
where  $ W_{ij}(t)  =  I(X_i \geq t) / \widehat{G}_C(t)$ and $\widehat{G}_C(\cdot)$ is the estimated survival function of the censoring (e.g., the Kaplan-Meier estimator). A cause-specific time-independent  BS is defined by
\begin{equation*}
 \widehat{\mbox{BS}}_{j} = \sum_{t=1}^d w_j(t) \widehat{\mbox{BS}}_{j}(t) 
\end{equation*}
and finally, a global BS is given by 
$\widehat{\mbox{BS}} = \sum_{j=1}^M v_j \widehat{\mbox{BS}}_{j}$.

\subsection{Predictions}
Once the  parameters of the model are estimated, predictions for test data are provided. Consider a test observation with baseline covariates $Z$, \pkg{PyDTS} provides the following useful predictions: 
\begin{enumerate}
    \item The overall survival at each time point $t$, $t=1,\ldots,d$, $$\widehat{S}(t|Z)=\prod_{k=1}^t\left\{1-\sum_{j=1}^M \widehat{\lambda}_{j}(k|Z)\right\} \, .$$
    \item The hazard of each failure type $j$ at each time point $t$, $j=1,\dots,M$, $t=1,\ldots,d$, 
$$
\widehat{\lambda}_j(t|Z)= \frac{\exp(\widehat{\alpha}_{jt}+Z^T\widehat{\beta}_j)}{1+\exp(\widehat{\alpha}_{jt}+Z^T\widehat{\beta}_j)} \, .
$$ 
    \item The probability of event type $j$ at each time $t$, $j=1,\dots,M$,
    $t=1,\ldots,d$, 
    $$
    \widehat{\Pr}(T=t,J=j|Z)=\widehat{\lambda}_j(t|Z) \widehat{S}(t-1|Z) \, .
    $$
    \item The cumulative incident function (CIF) of event type $j$ at any time $t$, $j=1,\dots,M$, $t=1,\ldots,d$,  
    $$\widehat{F}_j(t|Z)=\sum_{k=1}^t \widehat{\Pr}(T=k,J=j|Z) \, .$$
\end{enumerate}

\section{Simulating Discrete Time Survival Data with Competing Events}
\label{sec:datageneration}

\pkg{PyDTS} provides \texttt{EventTimesSampler} (ETS) class for sampling discrete-time survival data with competing risks and right censoring under the log-link model described  by  Equation~(\ref{eq:logis}). 

\subsection{Covariates}
\begin{figure} 
    \centering
    \includegraphics[width=\textwidth]{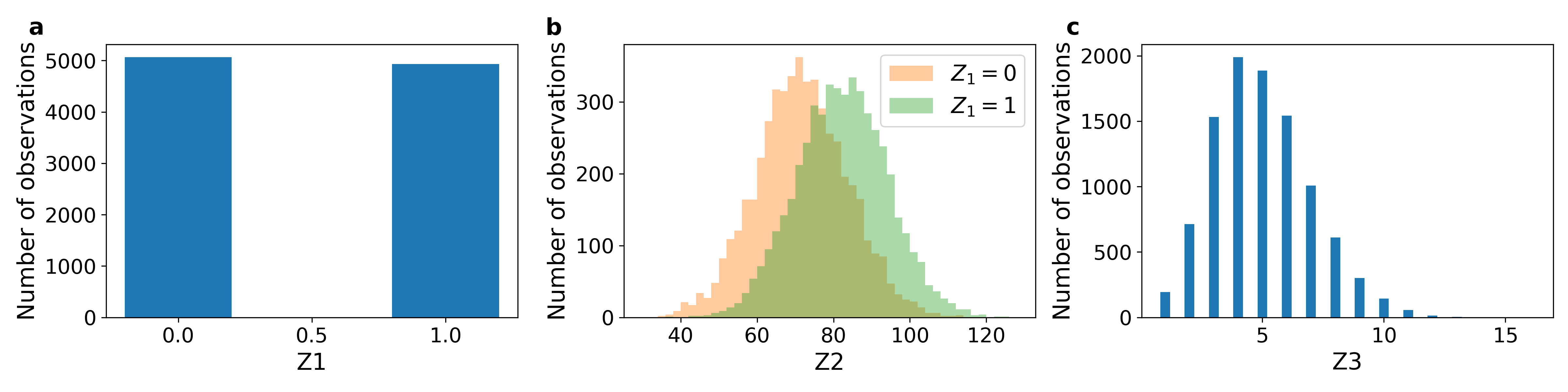}
    \caption{Sampled covariates. \textbf{a.} The observed distribution of $Z_1$, \textbf{b.} The observed distribution of $Z_2|Z_1$. \textbf{c.} The distribution of $Z_3$.}
    \label{fig:ets_observations}
\end{figure}

A user-supplied covariates should be passed to ETS. For example, consider a setting with $n=10,000$ independent observations and the following covariates
$$
Z_1 \sim  \mbox{Bernoulli(0.5)} 
$$
$$
Z_2 | Z_1 \sim \mbox{Normal}(72 + 10 Z_1, 12) \, ,
$$
and
$$
Z_3 \sim  1+\mbox{Poisson}(4) \, .
$$
Any sampling framework can be used for creating the covariates' dataframe. Here, we use \pkg{Numpy} \citep{harris2020array} by with the following lines of code
\begin{CodeInput}
observations_df = pd.DataFrame(columns=['Z1', 'Z2', 'Z3'])
observations_df['Z1'] = np.random.binomial(n=1, p=0.5, 
                                           size=n_observations)
observations_df.loc[observations_df.loc[observations_df['Z1'] == 0].index, 'Z2'] = 
    np.random.normal(loc=72, scale=12, 
                     size=n_observations-observations_df['Z1'].sum())
observations_df.loc[observations_df.loc[observations_df['Z1'] == 1].index, 'Z2'] = 
    np.random.normal(loc=82, scale=12, 
                     size=observations_df['Z1'].sum())
observations_df['Z3'] = 1 + np.random.poisson(lam=4, size=n_observations)
\end{CodeInput}
The sampled values are presented in Figure~\ref{fig:ets_observations}. 

\subsection{Event Times}
The ETS function assumes that the possible failure times are $1, \ldots, d$, and the user should supply the value of $d$. Clearly, the time intervals can be irregularly spaced and variable in size. For instance, discrete-time categories 1, 2, and 3 could correspond to specific days like Tuesday, Thursday, and Friday-Sunday, respectively. In the current example, we chose $d=7$. 

For the competing-events setting the user should decide on the number of competing events, and the values of model parameters, $\alpha_{jt}$, $\beta_j$, $t=1,\ldots,d$, $j=1,\ldots,M$. For example, consider $M=2$ competing events and
\begin{eqnarray*}
    &\alpha_{1t} = & -1 - 0.3 \log t \, , \, t =1, \ldots, 7 \\
    &\alpha_{2t} = & -1.75 - 0.15 \log t \, , \, t=1, \ldots, 7 \\
    &\beta^T_{1} = & (-\log 0.8, -\log 1.4, -\log 3) \\
    &\beta^T_{2} = & (-\log 1, -\log 0.95, -\log 2) \, .
\end{eqnarray*}
All together, the ETS function is defined for sampling event-type and event-time, and adding it to the data frame, as follows:
\begin{CodeInput}
ets = EventTimesSampler(d_times=7, j_event_types=2)
coefficients_dict = {
    "alpha": {
        1: lambda t: -1 - 0.3 * np.log(t),
        2: lambda t: -1.75 - 0.15 * np.log(t),
    },
    "beta": {
        1: -np.log([0.8, 1.4, 3]),
        2: -np.log([1, 0.95, 2]),
}}
observations_df = ets.sample_event_times(observations_df, coefficients_dict)
\end{CodeInput}

If the sampled covariates and parameters' values lead to impossible survival probabilities (i.e., negative or greater than one), the sampling process will be terminated with an error message. In such scenarios, it may be useful to adjust the coefficients or constrain extreme values of the covariates to ensure that the probabilities are appropriate and the sampling process is executed successfully.

\subsection{Censoring Time}
Two types of right censoring are implemented in \pkg{PyDTS}, administrative and random right censoring. For administrative censoring, $J_i=0$ and $T_i = d + 1$. These are the default values of observations for which the sampled event type was observed to be greater than $d$. Random right censoring is optional and could be either dependent or independent of the covariates. For example, assume 
$$
\Pr (C_i = t) = 0.05 \quad  t=1, \ldots, 7 \, .
$$
The censoring times can be sampled by
\begin{CodeInput}
prob_lof_at_t = [0.05, 0.05, 0.05, 0.05, 0.05, 0.05, 0.05]
observations_df = ets.sample_independent_lof_censoring(observations_df, prob_lof_at_t)
\end{CodeInput}
To generate right-censoring times that depend on the covariates, the user should supply to censoring hazard function, $\lambda_c(t|Z)$ in the form of   Equation~(\ref{eq:logis}). For example, 
\begin{CodeInput}
censoring_coef_dict = {
    "alpha": {
        0: lambda t: -0.3 - 0.3 * np.log(t),
    },
    "beta": {
        0: -np.log([8, 0.95, 6]),
}}
observations_df = ets.sample_hazard_lof_censoring(observations_df, 
        censoring_coef_dict)
\end{CodeInput}
Finally, the observed data should be updated by $X_i = min(T_i, C_i)$ and $J_i$ as follows
\begin{CodeInput}
observations_df = ets.update_event_or_lof(observations_df)
\end{CodeInput}
The first five observations of the sampled data are shown in Table \ref{tab:first_observations}.

\section{Examples}
\label{ref:illustations}

The classes \texttt{DataExpansionFitter} and \texttt{TwoStagesFitter} of
\pkg{PyDTS} implement the approach of \cite{lee_analysis_2018} and  the  two-step algorithm of \cite{meir_gorfine_dtsp_2023}, respectively.

\texttt{DataExpansionFitter} uses the \pkg{StatsModels} \citep{seabold2010statsmodels} package of \proglang{Python}, with the default GLM initiation. Other initial values can be supplied with \texttt{models\_kwargs} and \texttt{model\_fit\_kwargs} arguments of  \texttt{fit()}.

\texttt{TwoStagesFitter} uses \texttt{CoxPHFitter} of \pkg{Lifelines} \citep{davidson-pilon_lifelines_2019} for estimating $\beta_j$, $j=1,\ldots,M$, with  Efron's correction for ties \citep{efron_efficiency_1977}, implemented in \pkg{Lifelines}. For the estimation of $\alpha_{jt}$, \texttt{TwoStagesFitter} uses \texttt{minimize} of \pkg{Scipy} \citep{2020SciPy-NMeth} with $method=``BFGS''$. It uses the default values, except for the convergence tolerance that was set to be $gtol=1e-7$, which is more strict than the default ($1e-5$). 

Consider $M=2$ competing events,  $d=30$ discrete time points, $n=50,000$ observations and $p=5$ covariates sampled from a \mbox{Uniform}[0,1] distribution, $\alpha_{1t} = -1 -0.3 \log t $, $\alpha_{2t} = -1.75 -0.15 \log t$,
$$\beta^T_1 = -(\log 0.8, \log 3, \log 3, \log 2.5, \log 2)$$ 
and 
$$\beta^T_2 = -(\log 1, \log 3, \log 4, \log 3, \log 2) \, .$$ For the censoring distribution we employ additional available option that randomly assigns a subset of the sample to have uncensored observed time, utilizing the parameter $censoring\_prob$. As an example, if $censoring\_prob=0.8$, then the censoring time for 20\% of the sample is set to $T_{max}+1$, whereas for the remaining 80\% of the sample, censoring times are randomly sampled from a discrete uniform distribution $\mbox{Uniform}\{1,...,31\}$.
 A dataset based on these choices can be generated  by 
\begin{CodeInput}
real_coef_dict = {
    "alpha": {
        1: lambda t: -1 - 0.3 * np.log(t),
        2: lambda t: -1.75 - 0.15 * np.log(t)
    },
    "beta": {
        1: -np.log([0.8, 3, 3, 2.5, 2]),
        2: -np.log([1, 3, 4, 3, 2])
}}
patients_df = generate_quick_start_df(n_patients=50000, n_cov=5, 
                                      d_times=30, j_events=2, pid_col='pid', 
                                      seed=0, censoring_prob=0.8,
                                      real_coef_dict=real_coef_dict)
\end{CodeInput}

\begin{figure} 
    \centering
    \includegraphics[width=0.9\textwidth]{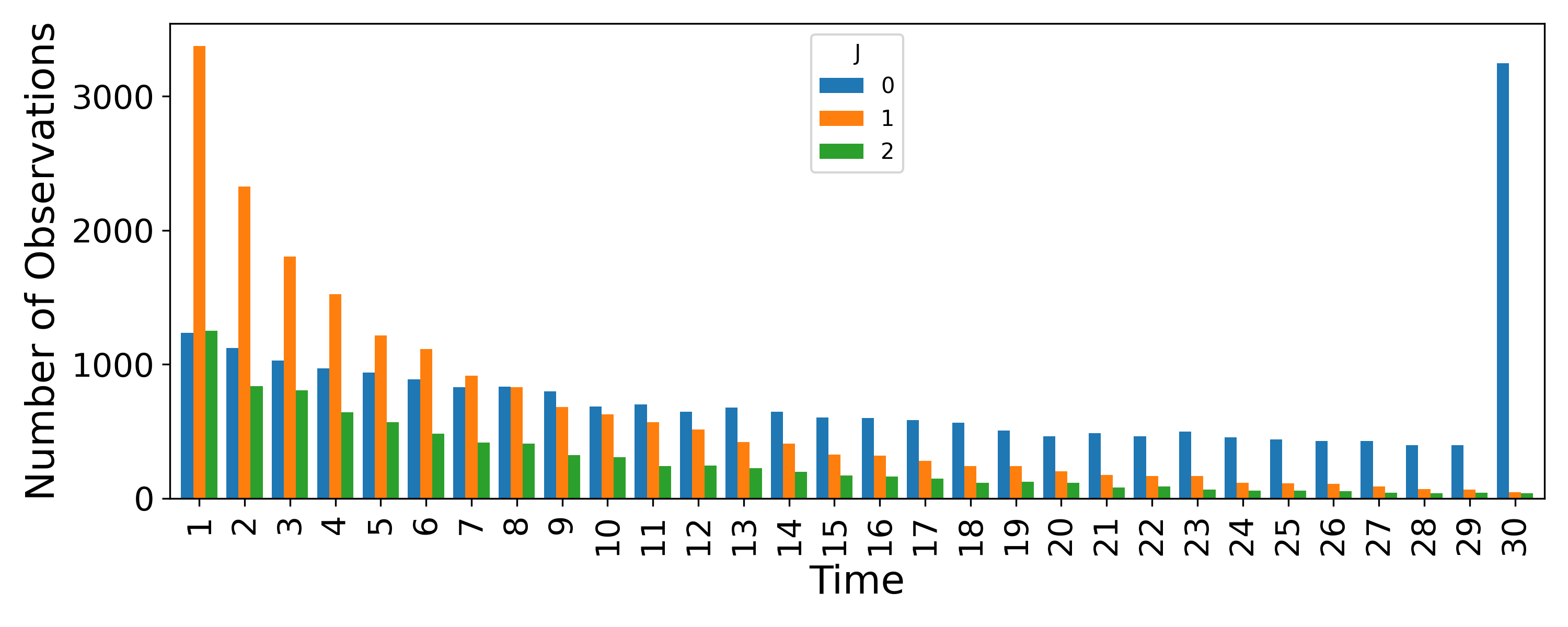}
    \caption{Events and censoring distribution of the simulated dataset with $n=50,000$ observations, $M=2$ event types, and $d=30$ time points.  $\alpha_{1t} = -1 -0.3 \log t$, $\alpha_{2t} = -1.75 -0.15\log(t)$, $\beta^T_1 = -(\log 0.8, \log 3, \log 3, \log 2.5, \log 2)$, 
$\beta^T_{2} = -(\log 1, \log 3, \log 4, \log 3, \log 2)$ and censoring times are sampled from $\mbox{Uniform}\{1,...,d+1\}$.}\label{fig:sim_events}
\end{figure}

The estimation techniques require a sufficient number of observed failures for each type of failure at each discrete time point. As such, the initial step is to verify if this condition is met by the available data. This could be done by
\begin{CodeInput}
patients_df.groupby(['J', 'X'])['pid'].count().unstack('J')
\end{CodeInput}
or by using the plot\_events\_occurrence() function available in \pkg{PyDTS} and visualizing the observed distributions, as demonstrated in Figure \ref{fig:sim_events}. Pre-processing suggestions when the data do not comply with this requirement will be presented in the Data Regrouping Example of Section \ref{subsec:dataprep}. 

The estimation procedure of \cite{meir_gorfine_dtsp_2023} can be applied by
\begin{CodeInput}
new_fitter = TwoStagesFitter()
new_fitter.fit(df=patients_df.drop(['C', 'T'], axis=1))
\end{CodeInput}
A summary table of the estimated coefficients with their standard errors can be generated by 
\begin{CodeInput}
new_fitter.print_summary()
\end{CodeInput}
The estimated standard errors of the estimates can be extracted by
\begin{CodeInput}
new_fitter.get_beta_SE()
\end{CodeInput}
which results in the following output
\begin{CodeOutput}
           j1_params     j1_SE  j2_params     j2_SE
covariate                                          
Z1          0.187949  0.025068   0.040169  0.037807
Z2         -1.100792  0.025610  -1.100246  0.038696
Z3         -1.093466  0.025726  -1.410202  0.039280
Z4         -0.874521  0.025437  -1.097849  0.038642
Z5         -0.652655  0.025280  -0.654501  0.038179
\end{CodeOutput}
Additionally, it is possible to visualize the estimated coefficients using
\begin{CodeInput}
new_fitter.plot_all_events_alpha()
new_fitter.plot_all_events_beta()
\end{CodeInput}

\begin{figure}
    \centering
    \includegraphics[width=\textwidth,height=\textheight,keepaspectratio]{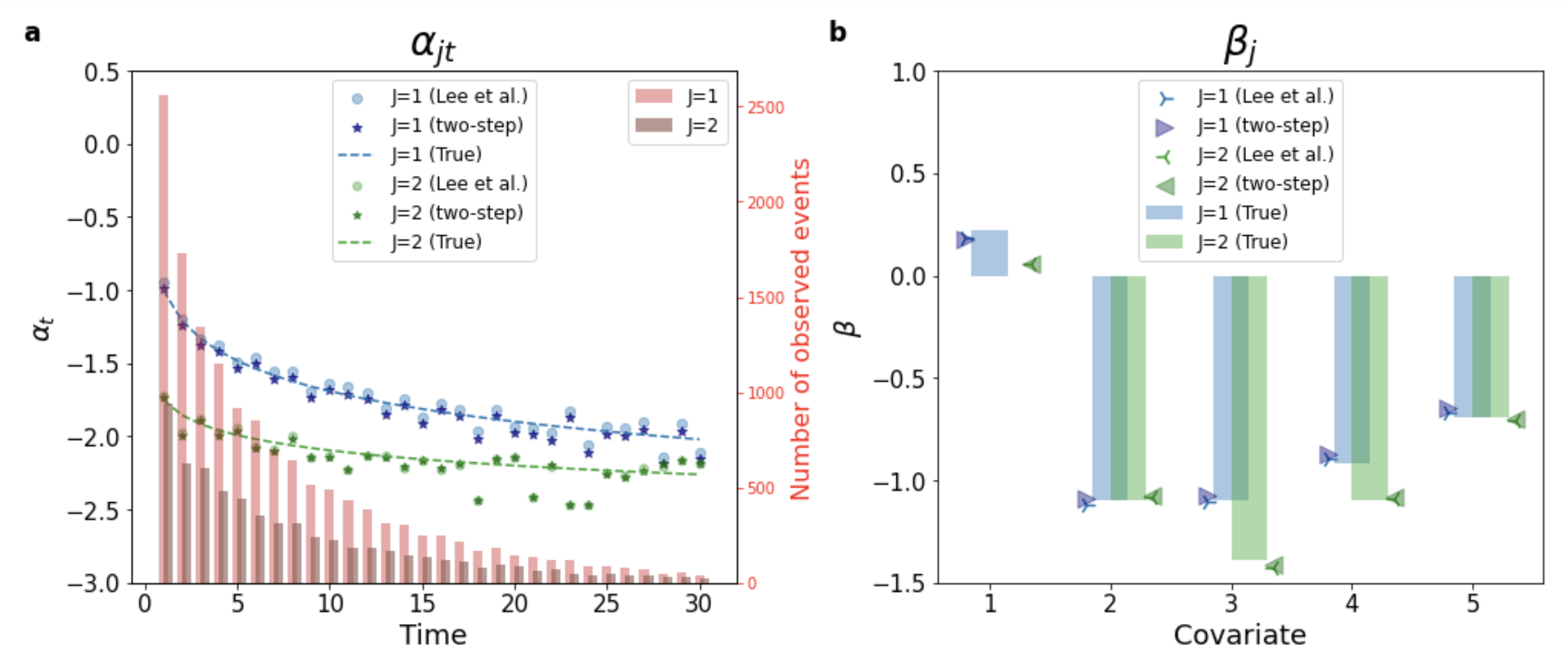}
    \caption{Results of one simulated dataset, $n=50,000$: Estimation results of \cite{lee_analysis_2018} and the two-step procedure of \cite{meir_gorfine_dtsp_2023} along with the true parameters' values. \textbf{a:} Results of $\alpha_{jt}$. True values of $\alpha_{1t}$ are in dashed blue line, estimates of Lee et al. and the two-step algorithm  are in light blue circles and blue stars, respectively. Similarly, true values and estimates of $\alpha_{2t}$ are in green scale. Number of observed events at each time $t$ is shown in red and brown bars. \textbf{b:} Results of $\beta_{j}$. True values of $\beta_{1}$ are in light blue bars, estimates of Lee et al. and the two-step algorithm are in light blue right arrowhead and blue right triangle, respectively. Similarly, true values and estimates of $\beta_{2}$ are in green scale.}\label{fig:single_run_comparison}
\end{figure}

The estimation method of \cite{lee_analysis_2018}  is performed by 
\begin{CodeInput}
fitter = DataExpansionFitter()
fitter.fit(df=patients_df.drop(['C', 'T'], axis=1))
fitter.print_summary()
\end{CodeInput}
The results are presented in Figure \ref{fig:single_run_comparison} and  Table \ref{tab:beta_comparison}. 

\begin{figure}
    \centering
    \includegraphics[width=\textwidth]{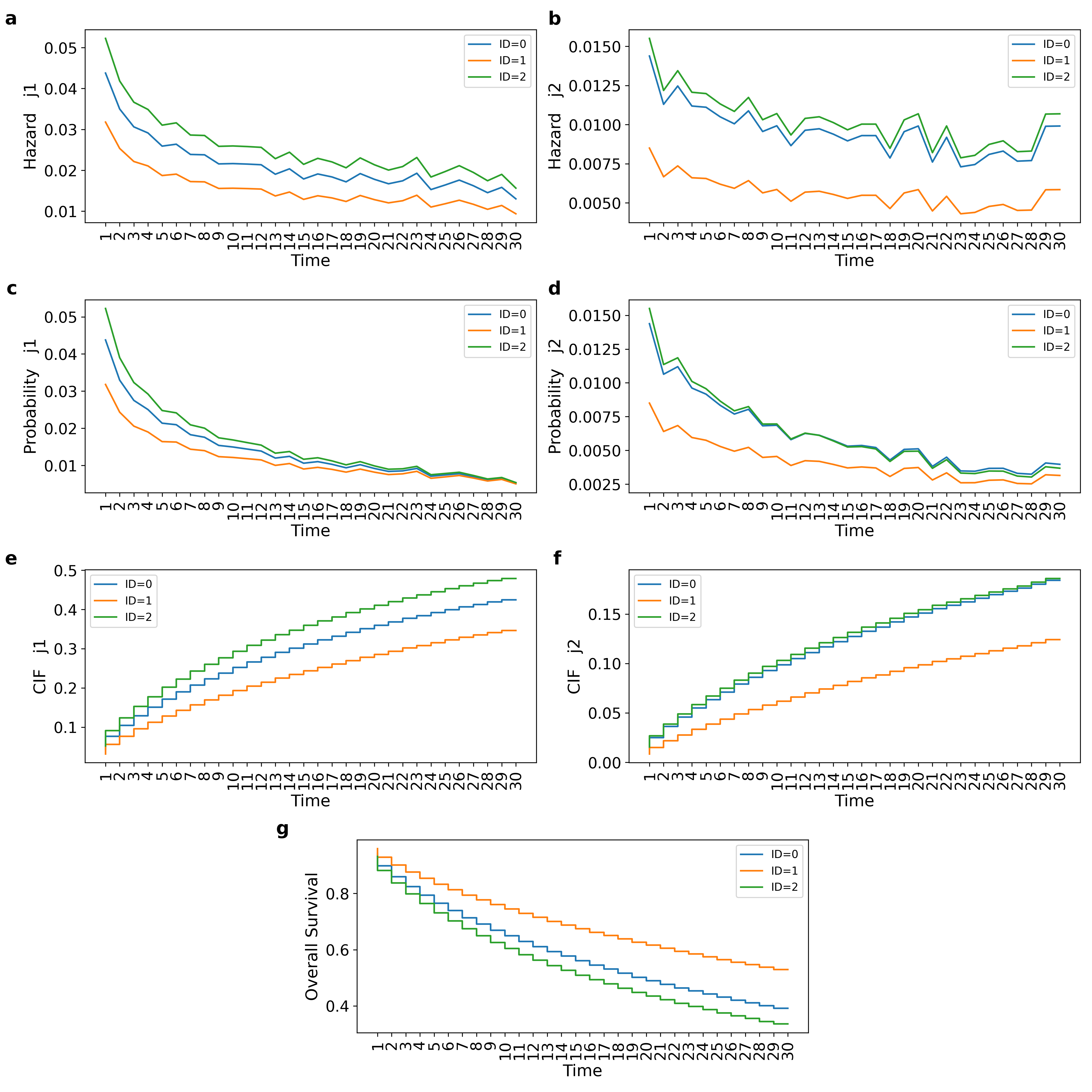}
    \caption{Predictions for three individuals of the test dataset: ID=0 (blue), ID=1 (orange), and ID=2 (green). \textbf{a.} Estimated hazard function of event type $j=1$, $\widehat{\lambda}_{1}(t)$. \textbf{b.} Estimated hazard function of event type $j=2$, $\widehat{\lambda}_{1}(t|Z)$. \textbf{c.} Estimated probability of event type $j=1$, $\widehat{\Pr}(T=t,J=1|Z)$. \textbf{d.} Estimated probability for event type $j=2$, $\widehat{\Pr}(T=t,J=2|Z)$. \textbf{e.} Estimated CIF of event type $j=1$, $\widehat{F}_1(t|Z)$. \textbf{f.} Estimated CIF of event type $j=2$, $\widehat{F}_2(t|Z)$. \textbf{g.} Estimated overall survival function, $\widehat{S}(t|Z)$. }\label{fig:pred_df}
\end{figure}

For generating the predictions of Section 2.7 for new observations  an input data frame pandas.DataFrame() is required, containing for each observation a vector of baseline covariates in the same order used in fit(). 
The following code provides predictions for the first three individuals in the dataset $patients\_df$
\begin{CodeInput}
pred_df = new_fitter.predict_cumulative_incident_function(
                patients_df.drop(['J', 'T', 'C', 'X'], axis=1).head(3))
print(pred_df)
\end{CodeInput}
The results are given in a tabular form. The following output shows  $\widehat{S}(t|Z)$ (overall survival), $\widehat{\lambda}_j(t|Z)$ (hazard functions), $\widehat{\Pr}(T=t,J=j|Z)$ (joint distribution of $T,J$)  and $\widehat{F}_j(t|Z)$ (CIFs). Figure \ref{fig:pred_df} presents the entire curves of these three individuals.

\begin{CodeOutput}
                          ID=0      ID=1      ID=2
Z1                    0.548814  0.645894  0.791725
Z2                    0.715189  0.437587  0.528895
Z3                    0.602763  0.891773  0.568045
Z4                    0.544883  0.963663  0.925597
Z5                    0.423655  0.383442  0.071036
overall_survival_t1   0.942684  0.960628  0.932938
overall_survival_t2   0.899636  0.930545  0.883002
    ...                 ...        ...      ...
overall_survival_t29  0.406209  0.545669  0.348543
overall_survival_t30  0.397051  0.537568  0.339489
hazard_j1_t1          0.043097  0.031017  0.051717
hazard_j1_t2          0.034478  0.024750  0.041448
    ...                 ...        ...      ...
hazard_j1_t29         0.015702  0.011211  0.018951
hazard_j1_t30         0.012799  0.009130  0.015457
hazard_j2_t1          0.014218  0.008355  0.015345
hazard_j2_t2          0.011188  0.006566  0.012077
    ...                 ...        ...      ...
hazard_j2_t29         0.009730  0.005707  0.010505
hazard_j2_t30         0.009745  0.005716  0.010521
prob_j1_at_t1         0.043097  0.031017  0.051717
prob_j1_at_t2         0.032501  0.023776  0.038668
    ...                 ...        ...      ...
prob_j1_at_t29        0.006545  0.006223  0.006806
prob_j1_at_t30        0.005199  0.004982  0.005387
prob_j2_at_t1         0.014218  0.008355  0.015345
prob_j2_at_t2         0.010546  0.006308  0.011267
    ...                 ...        ...      ...
prob_j2_at_t29        0.004056  0.003168  0.003773
prob_j2_at_t30        0.003958  0.003119  0.003667
cif_j1_at_t1          0.043097  0.031017  0.051717
cif_j1_at_t2          0.075599  0.054792  0.090385
    ...                 ...        ...      ...
cif_j1_at_t29         0.415879  0.335485  0.471603
cif_j1_at_t30         0.421078  0.340468  0.476990
cif_j2_at_t1          0.014218  0.008355  0.015345
cif_j2_at_t2          0.024765  0.014663  0.026612
    ...                 ...        ...      ...
cif_j2_at_t29         0.177912  0.118845  0.179853
cif_j2_at_t30         0.181870  0.121964  0.183520
\end{CodeOutput}

\begin{figure}
    \centering
    \includegraphics[width=\textwidth,height=\textheight,keepaspectratio]{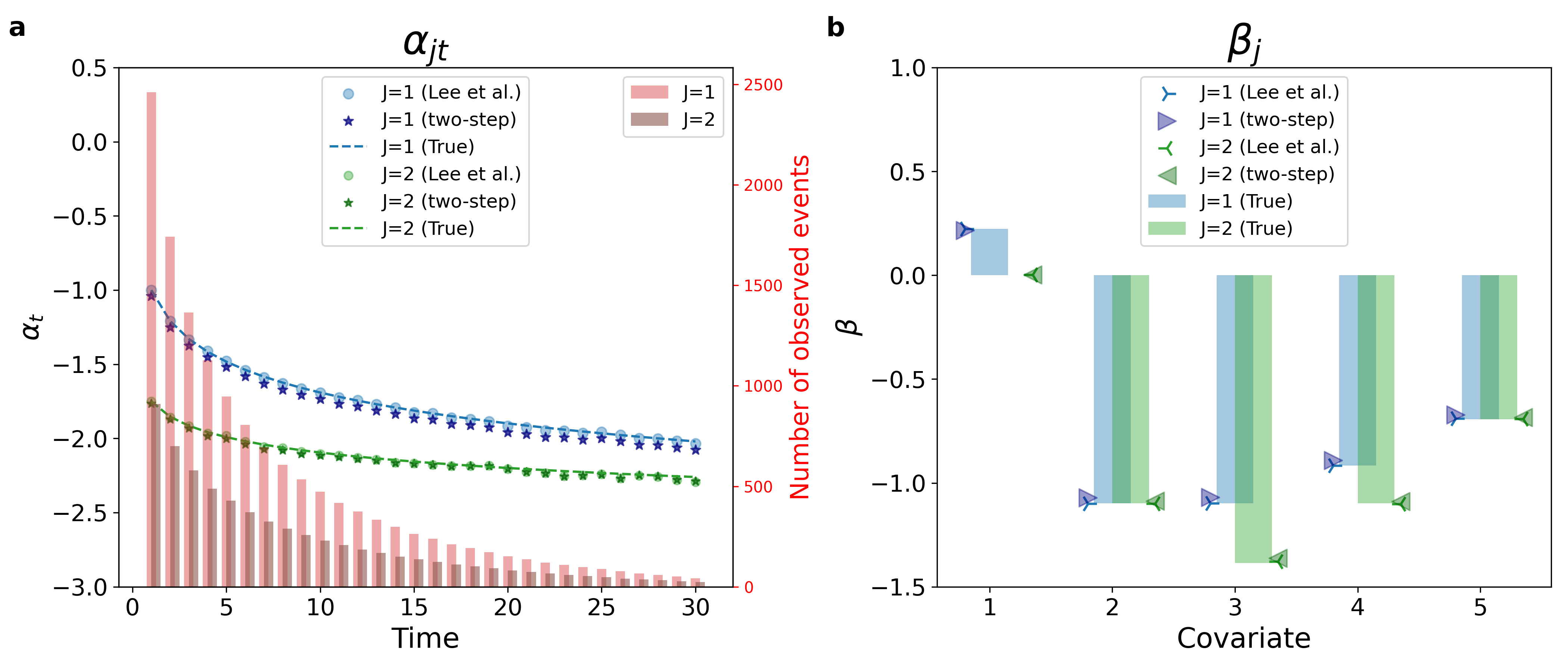}
    \caption{Simulation results of \cite{lee_analysis_2018}, the two-step algorithm of \cite{meir_gorfine_dtsp_2023} the true parameters' values based on 100 repetitions. \textbf{a:} Results of $\alpha_{jt}$. True values of $\alpha_{1t}$ are in dashed blue line, estimates of Lee et al. and the two-step algorithm are in light blue circles and blue stars, respectively. Similarly, true values and estimates of $\alpha_{2t}$ are in green scale. Number of observed events at each time $t$ is shown in red and brown bars. \textbf{b:} Results of $\beta_{j}$. True values of $\beta_{1}$ are in light blue bars, estimates of Lee et al. and the two-step algorithm are in light blue right arrowhead and blue right triangle, respectively. Similarly, true values and estimates of $\beta_{2}$ are in green scale. }\label{fig:models_params}
\end{figure}

\begin{figure} 
    \centering
    \includegraphics[width=\textwidth, keepaspectratio]{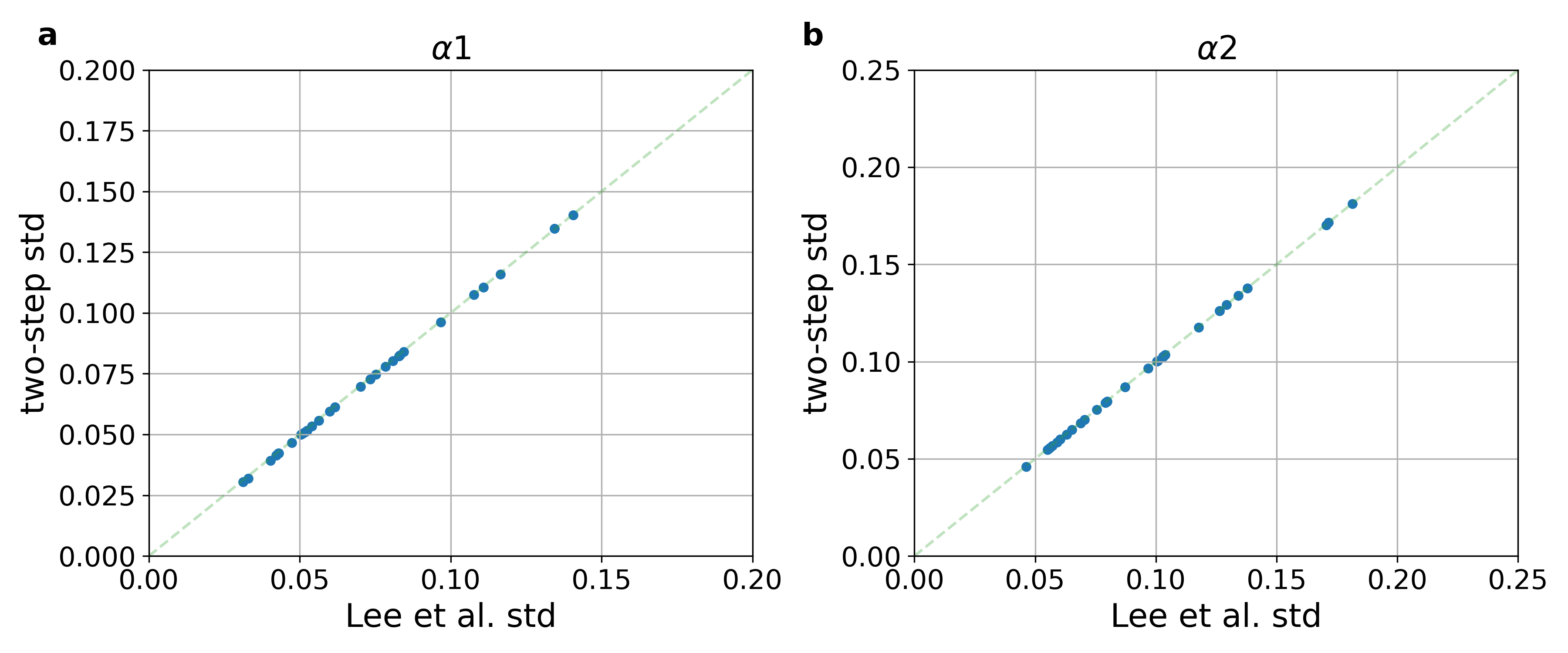}
    \caption{Simulation results based on 100 repetitions: standard errors of Lee et al. versus that of the two-step algorithm of \cite{meir_gorfine_dtsp_2023}.  \textbf{a:} Standard errors of $\widehat{\alpha}_{1t}$. \textbf{b:} Standard errors of $\widehat{\alpha}_{2t}$.}\label{fig:std}
\end{figure}

\begin{figure} 
    \centering
    \includegraphics[width=0.7\textwidth,height=\textheight,keepaspectratio]{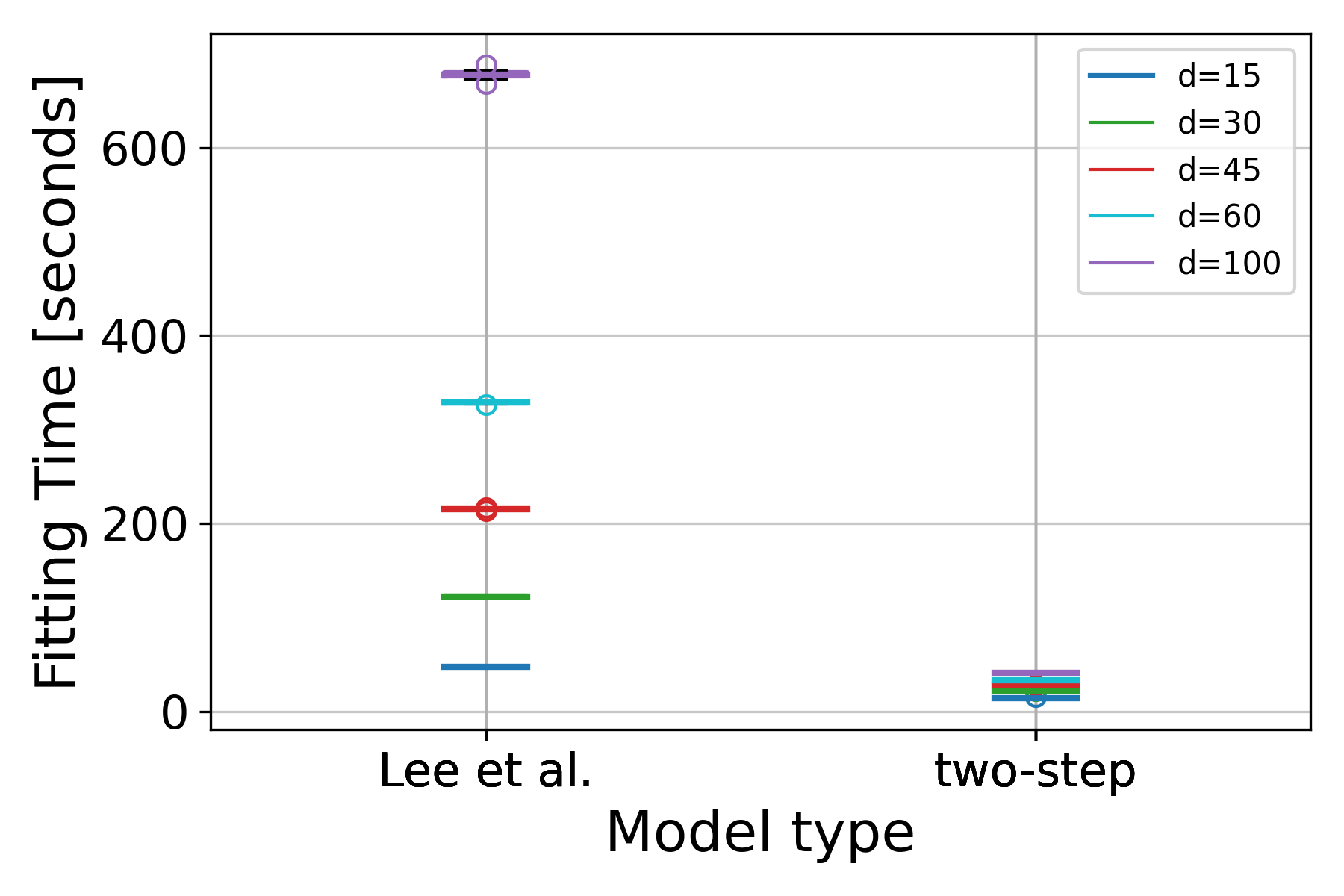}
    \caption{Simulation results based on 10 repetitions for each value of $d$: box-plots of computation times of Lee et al. and the two-step algorithm.}\label{fig:timing}
\end{figure}

We conducted a small simulation study demonstrating and comparing the two estimation procedures. The simulation is based on the data generation described above with 100 repetitions. The results are summarized in Figures \ref{fig:models_params}--\ref{fig:timing} and Table \ref{tab:rep_beta_comparison}. Evidently, both methods  provide similar point estimators and standard errors, as expected. Moreover, both methods perform very well in terms of bias (Figures ~\ref{fig:models_params} and \ref{fig:std}). However, as evident by Figure~\ref{fig:timing}, the computation time of the two-step approach is shorter depending on $d$, and the improvement factor increases as a function of $d$. For additional simulation results, the reader is referred to \cite{meir_gorfine_dtsp_2023}.

\subsection{Regularization}
Regularized regression can be easily accommodated only with \texttt{TwoStagesFitter} where we first estimate $\beta_j$ and then $\alpha_{jt}$. Regularization is introduced  by  CoxPHFitter \citep{davidson-pilon_lifelines_2019} with  event-specific tuning parameters, $\eta_j \geq 0$, and $l1\_ratio$ argument. For each $j$, usually, a path of models in $\eta_j$ are fitted, and the value of $l1\_ratio$ defines the type of prediction model. In particular,  ridge regression \citep{hoel1970ridge} is performed by setting  $l1\_ratio=0$,   lasso \citep{tibshirani1996regression} by $l1\_ratio=1$, and elastic net \citep{zou2005regularization} by $0 < l1\_ratio <1$.

For example, lasso regression with two competing events, $\eta_1=0.003$ and $\eta_2=0.005$, can be applied by
\begin{CodeInput}
L1_regularized_fitter = TwoStagesFitter()
fit_beta_kwargs = {
    'model_kwargs': {
        1: {'penalizer': 0.003, 'l1_ratio': 1},
        2: {'penalizer': 0.005, 'l1_ratio': 1}
}}
L1_regularized_fitter.fit(df=patients_df.drop(['C', 'T'], axis=1),
                          fit_beta_kwargs=fit_beta_kwargs)
\end{CodeInput}

Regularization methods can be applied also with different penalty values for each covariates by passing a vector of length $p$ (instead of a scalar) to ``penalizer''. This could be useful, for example, when some covariates should not be penalized. For example, lasso regression with penalty for the first four covariates and leaving the last one non-penalized  can be done by
\begin{CodeInput}
L1_regularized_fitter = TwoStagesFitter()
fit_beta_kwargs = {
    'model_kwargs': {
        1: {'penalizer': np.array([0.01, 0.01, 0.01, 0.01, 0]), 'l1_ratio': 1},
        2: {'penalizer': np.array([0.05, 0.05, 0.05, 0.05, 0]), 'l1_ratio': 1}
}}
L1_regularized_fitter.fit(df=patients_df.drop(['C', 'T'], axis=1),
                          fit_beta_kwargs=fit_beta_kwargs)
\end{CodeInput}

In penalized regression, one should fit a path of models in each $\eta_j$, $j=1,\ldots,M$. The final set of values of $\eta_1,\ldots,\eta_M$ corresponds to the values yielding the best results in terms of  pre-specified criteria, such as  maximizing $\widehat{\mbox{AUC}}_j$ and  $\widehat{\mbox{AUC}}$, or minimizing $\widehat{\mbox{BS}}_j$ and $\widehat{\mbox{BS}}$. The default criteria in \pkg{PyDTS} is  maximizing the global AUC, $\widehat{\mbox{AUC}}$. Two $M$-dimensional grid search options are implemented, \texttt{PenaltyGridSearch} when the user provides train and test datasets, and \texttt{PenaltyGridSearchCV} for applying a K-fold cross validation (CV) approach. 

By executing 
\begin{CodeInput}
penalizers = np.exp([-2, -3, -4, -5, -6])
grid_search = PenaltyGridSearch()
optimal_set = grid_search.evaluate(train_df, test_df, l1_ratio=1, 
                                   penalizers=penalizers,
                                   metrics=['IBS', 'GBS', 'IAUC', 'GAUC']) 
\end{CodeInput}
all the four optimization criteria are calculated over the $M$-dimensional grid and $optimal\_set$ includes the optimal values of $\eta_1,\ldots,\eta_M$ based on $\widehat{\mbox{AUC}}$. Here, the optimal set based on $\widehat{\mbox{AUC}}$ is $\log\eta_1=-6$ and $\log\eta_2=-6$. The user can choose the set of $\eta_j$, $j=1,\ldots,M$, values that optimizes other desired criteria. For example, the set that minimizes $\widehat{\mbox{BS}}$ can be selected as follows
\begin{CodeInput}
print(grid_search.convert_results_dict_to_df(grid_search.global_bs).idxmin())
\end{CodeInput}
which results in the optimal set $\log\eta_1=-6$ and $\log\eta_2=-3$.

For applying the two-stage regularized regression under a specific set of $\eta_j$, for example $optimal\_set$, use
\begin{CodeInput}
optimal_two_stages_fitter = grid_search.get_mixed_two_stages_fitter(
                                optimal_set) 
\end{CodeInput}

Alternatively, 5-fold CV is performed by
\begin{CodeInput}
penalizers = np.exp([-2, -3, -4, -5, -6])
grid_search_cv = PenaltyGridSearchCV()
results_df = grid_search_cv.cross_validate(patients_df, l1_ratio=1, 
                                penalizers=penalizers, n_splits=5, 
                                metrics=['IBS', 'GBS', 'IAUC', 'GAUC'])
optimal_set = results_df['Mean'].idxmax()
print(results_df)
\end{CodeInput}
The number of folds is defined by $n\_splits$. The output is a tabular data  of type pandas.DataFrame with the mean and SE of the fold-wise $\widehat{\mbox{AUC}}$, for example 
\begin{CodeOutput}
                           Mean        SE
log(eta_1) log(eta_2)                    
-2.0       -2.0        0.638629  0.004971
           -3.0        0.639105  0.004849
           -4.0        0.639125  0.004856
           -5.0        0.637303  0.005168
           -6.0        0.488880  0.005754
-3.0       -2.0        0.638909  0.004880
           -3.0        0.638909  0.004847
           -4.0        0.639026  0.004849
           -5.0        0.637421  0.005146
           -6.0        0.488880  0.005753
-4.0       -2.0        0.638782  0.004862
           -3.0        0.638986  0.004814
           -4.0        0.639086  0.004812
           -5.0        0.637501  0.005085
           -6.0        0.488879  0.005752
-5.0       -2.0        0.563814  0.006279
           -3.0        0.563814  0.006280
           -4.0        0.563814  0.006282
           -5.0        0.563766  0.006334
           -6.0        0.614951  0.003800
-6.0       -2.0        0.568809  0.006638
           -3.0        0.568809  0.006638
           -4.0        0.568809  0.006643
           -5.0        0.568761  0.006725
           -6.0        0.630466  0.005484
\end{CodeOutput}
and the other specified metrics are available as attributes of the \texttt{PenaltyGridSearchCV} object.

\subsection{Performance Measures}
Model evaluation on test data or by CV,  can be done using the evaluation functions available in \pkg{PyDTS} and the measures of performance presented in Section \ref{sec:performance}.   

For example, in the following code, the survival models are estimated based on the two-stage approach and the dataset $train\_df$. Assume that the event of main interest is $j=1$. Then,  $\pi_{i1}(t)$ are calculated and stored in $pred\_df$, and finally
 $\widehat{\mbox{AUC}}_1(t)$, $t=1,\ldots,d$, are provided by
\begin{CodeInput}
fitter = TwoStagesFitter()
fitter.fit(df=train_df, axis=1))
pred_df = fitter.predict_prob_event_j_all(test_df, event=1)
auc_1 = event_specific_auc_at_t_all(pred_df, event=1)
\end{CodeInput}
Other measures such as $\widehat{\mbox{AUC}}_1$, $\widehat{\mbox{BS}}_1$, $\widehat{\mbox{AUC}}$, and $\widehat{\mbox{BS}}$ can be calculated by
\begin{CodeInput}
pred_df = fitter.predict_prob_events(test_df)
ibs_1 = event_specific_integrated_brier_score(pred_df, event=1)
iauc_1 = event_specific_integrated_auc(pred_df, event=1)
bs = global_brier_score(pred_df)
auc = global_auc(pred_df)
\end{CodeInput}

Model evaluation based on K-fold CV and  \texttt{TwoStagesFitter} can be done by
\begin{CodeInput}
cross_validator = TwoStagesCV()
cross_validator.cross_validate(full_df=patients_df.drop(['C', 'T'], axis=1), 
                               n_splits=5, seed=0,
                               metrics=['BS', 'IBS', 'GBS', 
                                        'AUC', 'IAUC', 'GAUC'])
\end{CodeInput}
The user should provide the list of desired performance measures.

\subsection{Data Regrouping}
\label{subsec:dataprep}
\begin{figure}
    \centering
    \includegraphics[width=0.9\textwidth]{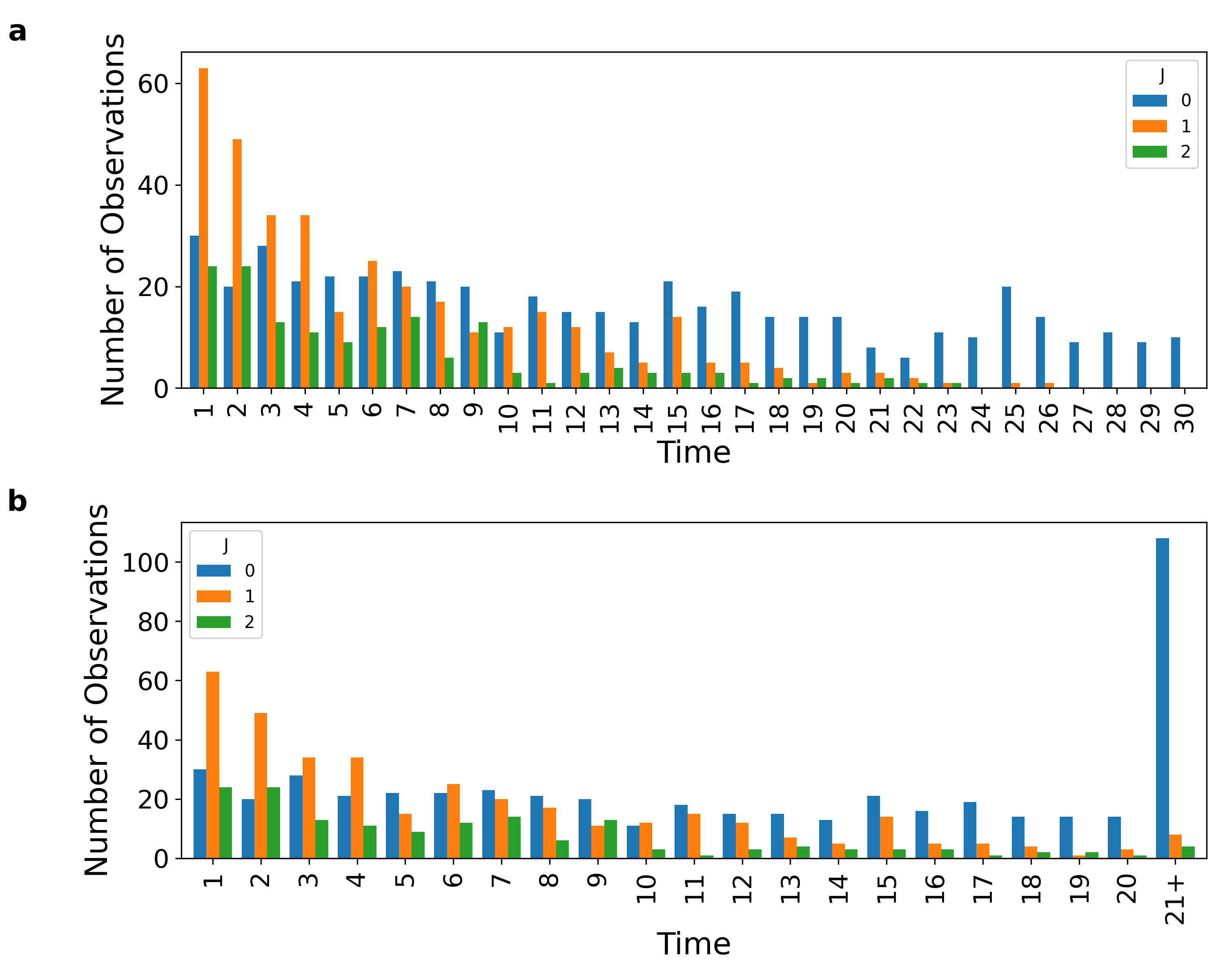}
    \caption{A simulated dataset: $n=1,000$ observations, $M=2$ failure type, $\alpha_{1t} = -1 -0.3 \log t$, $\alpha_{2t} = -1.75 -0.15\log t$, $\beta^T_1 = -(\log 0.8, \log 3, \log 3, \log 2.5, \log 2)$, 
$\beta^T_{2} = -(\log 1, \log 3, \log 4, \log 3, \log 2)$ and  $C_i \sim \mbox{Uniform}\{1,...,d+1\}$. \textbf{a.} Original data with $d=30$ possible event occurrence times. \textbf{b.} Regrouped data  - days 21 and beyond were combined into one category of time named $21+$. }\label{fig:events_tail}
\end{figure}

As previously mentioned, both estimation techniques are sensitive to the number of observed failures at each $(j,t)$, $j=1,\ldots,M$, $t=1,\ldots,d$.   
Consider the above simulation setting but with a smaller sample size of $n=1,000$ independent observations.  The number of events and censored observations at each time point is shown in Figure \ref{fig:events_tail}a. Evidently, there are too few observed events in later times,  one event was observed at $(j,t)=(1,25)$ and 0 for $(j,t)=(2,25)$.  Fitting model (\ref{eq:logis})  with this dataset would fail and result with an error message. One straightforward solution is to group adjacent time points towards the end of the distribution together, allowing for a sufficient accumulation of observed events for each type of event.
For this particular case, events that are observed on day 21 or later are classified into the time category labeled ``21+'' by
\begin{CodeInput}
df['X'].clip(upper=21, inplace=True)
\end{CodeInput}
The regrouped dataset, described in Figure~\ref{fig:events_tail}b, can successfully be used for estimating the survival models for $j=1,2$ and $t=1,\ldots,20,21+$. Regrouping neighboring time points is a useful solution in datasets with too few events at any time point.

\section{Case Study}
\label{sec:mimic-los}
The utility of \pkg{PyDTS} is demonstrated by analysing patients' length of stay (LOS)  in healthcare facilities. The analysis is similar to that of \cite{meir_gorfine_dtsp_2023}, based on the publicly accessible dataset of the Medical Information Mart for Intensive Care (MIMIC) - IV (version 2.0) \citep{johnson_mimic-iv_2022, goldberger_physiobank_2000}. 

Our goal is developing a survival model for predicting LOS in ICU based on patients’ characteristics upon arrival in intensive care unit (ICU). The study includes 25,170 ICU admissions that occurred between 2014 and 2020, with LOS ranging from 1 to 28 days, leading to multiple tied events at each time point. Three competing events observed in data: discharge to home ($J=1$, 69.0\%), transfer to another medical facility ($J=2$, 21.4\%), and in-hospital death ($J=3$, 6.1\%). Patients who left the ICU against medical advice (1.0\%) were considered censored, and administrative censoring was imposed for patients hospitalized for more than 28 days (2.5\%). The analysis includes 36 covariates in total, for each patient. For a detailed description of the data, the reader is referred to Section 4 of \cite{meir_gorfine_dtsp_2023}.

\begin{figure}[h!]
    \centering
    \includegraphics[width=0.8\textwidth]{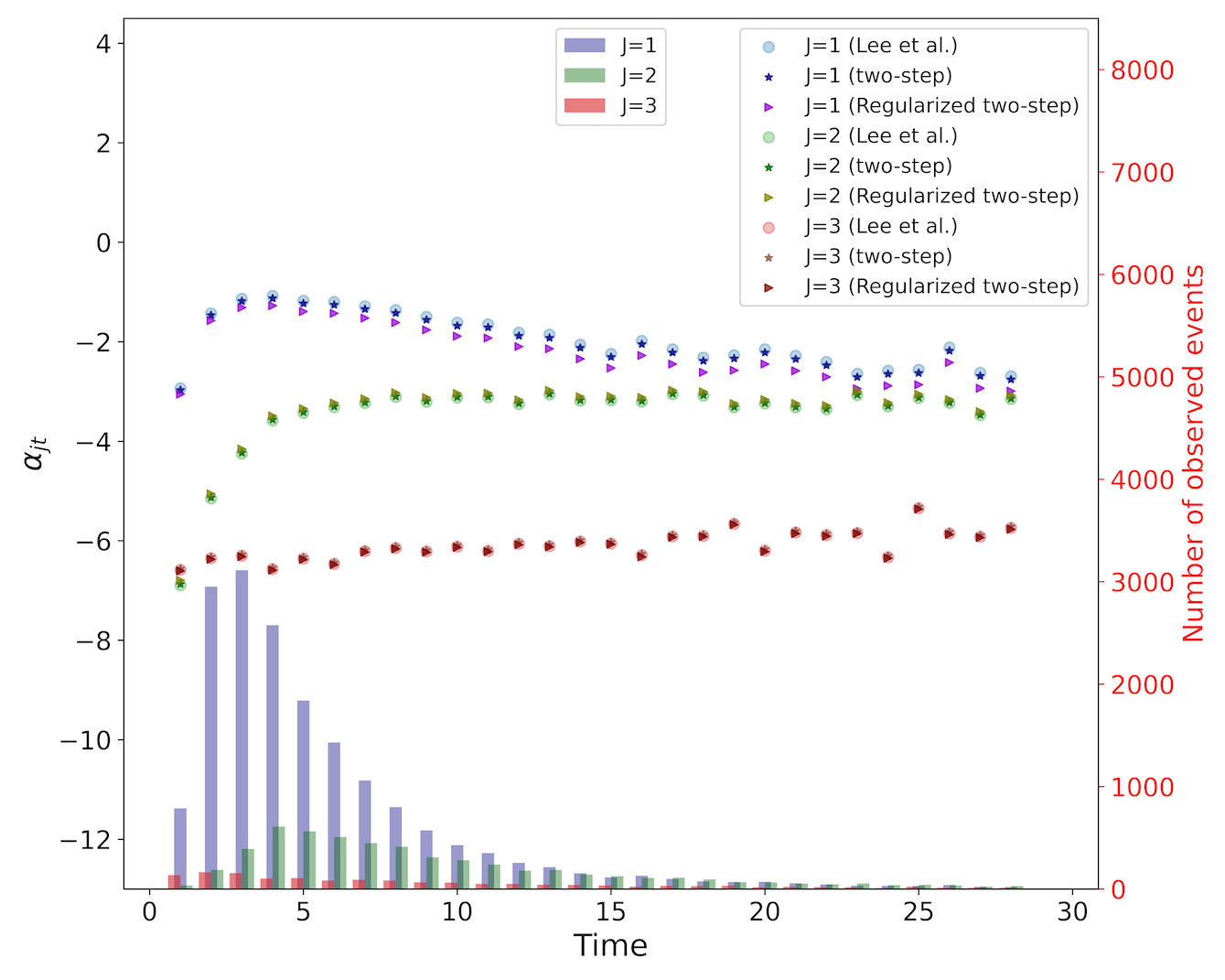}
    \caption{MIMIC dataset - LOS analysis. Results of estimated $\alpha_{jt}$ by the method of Lee et al. (circle), the  two-step approach of \cite{meir_gorfine_dtsp_2023} with no regularization (star) and  with lasso (left triangular). Numbers of observed events are shown in blue bars for home discharge ($j=1$), in green bars for further treatment ($j=2$), and in red bars for in-hospital death ($j=3$). Lasso estimates are based on  $\log \eta_1=-5$, $\log \eta_2=-9$ and $\log \eta_3=-11$.}
    \label{fig:mimic_alpha}
\end{figure}

\begin{figure}[h!]
    \centering
    \includegraphics[width=\textwidth]{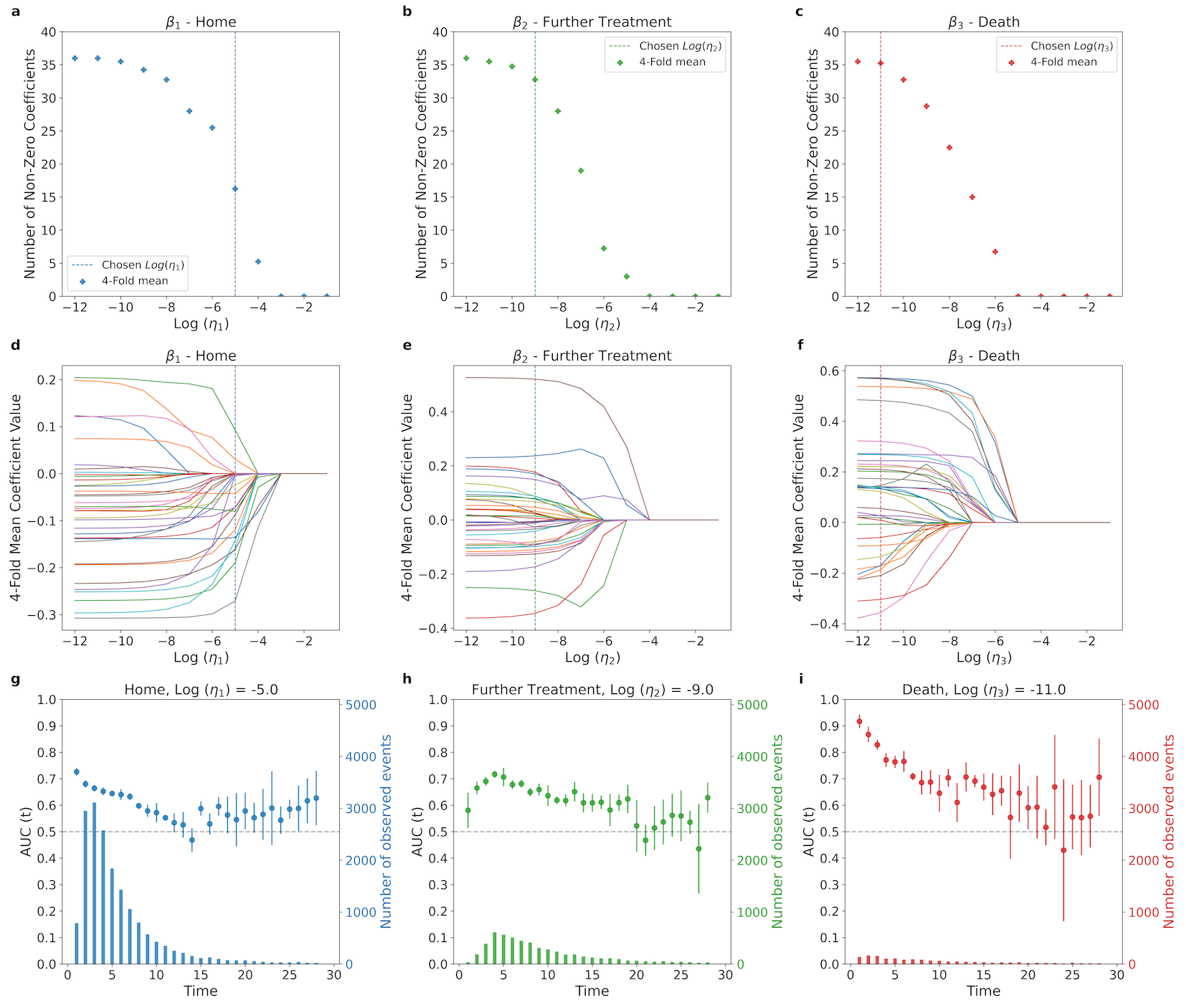}
    \caption{MIMIC dataset - LOS analysis. Regularized regression with 4-fold CV. The selected values of $\eta_j$ are shown in dashed-dotted lines on panels \textbf{a-f}. \textbf{a-c.} Number of non-zero coefficients for $j=1,2,3$. \textbf{d-f.} The estimated coefficients, as a function of $\eta_j$, $j=1,2,3$. \textbf{g-i.} Mean (and SD bars) of the 4 folds $\widehat{\mbox{AUC}}_j(t)$, $j=1,2,3$,  for the selected values  $\log \eta_1=-5$, $\log \eta_2=-9$ and $\log \eta_3=-11$. The number of observed events of each type is shown by bars.  
    }
    \label{fig:reg_mimic_results}
\end{figure}

Three estimation procedures were considered, \cite{lee_analysis_2018}, \cite{meir_gorfine_dtsp_2023} without regularization, and \cite{meir_gorfine_dtsp_2023} with lasso.  Lasso regression was performed with the best set of the tuning parameters $\eta_j$, $j=1,2,3$, based on the global AUC and 4-fold CV.  The following is the relevant code:
\begin{CodeInput}
lee_fitter = DataExpansionFitter()
lee_fitter.fit(df=patients_df)
two_step_fitter = TwoStagesFitter()
two_step_fitter.fit(df=patients_df)

penalizers = np.exp(np.arange(-12, -0.9, step=1))
penalty_cv_search = PenaltyGridSearchCV()
gauc_cv_results = penalty_cv_search.cross_validate(full_df=patients_df, l1_ratio=1,
    penalizers=penalizers,  n_splits=4)
chosen_set = gauc_cv_results['Mean'].idxmax()
\end{CodeInput}
The results are shown in Tables \ref{table:los-j1}-\ref{table:los-j3} and Figures \ref{fig:mimic_alpha}-\ref{fig:reg_mimic_results}.  Evidently, the estimated values of $\alpha_{jt}$ and $\beta_{j}$ based on \cite{lee_analysis_2018} and \cite{meir_gorfine_dtsp_2023} without regularization are similar. Figure \ref{fig:reg_mimic_results} shows the number of non-zero coefficients, the lasso regularization path, and $\widehat{\mbox{AUC}}_j(t)$ for each event type.  The selected values of $\log \eta_j$, $j=1,2,3$, were -5, -9 and -11, respectively. For the first 14 days of hospitalization $\widehat{\mbox{AUC}}_j(t)$ were higher than that of later times. This may be due to several reasons. Firstly, the number of observed events at early times are higher. Second, short LOS can be a consequence of the severity of illness, with short-term in-hospital death occurring for severe cases and short-term discharge for mild cases, making them easier to identify.  Lastly, as treatment progresses, the effect of the initial condition may decreases while the treatment effect increases, making it difficult to distinguish between events occurring during later times  based on the covariates measured upon admission.  The integrated cause-specific AUCs were $\widehat{\mbox{AUC}}_1=0.642$ (SD=0.002) (the SDs in parentheses are based on the 4 folds and do not take into account the variability due to the model estimation), $\widehat{\mbox{AUC}}_2=0.655$ (SD=0.012), and $\widehat{\mbox{AUC}}_3=0.740$ (SD=0.006), with a global $\widehat{\mbox{AUC}}=0.651$ (SD=0.003).  The integrated cause-specific BS were $\widehat{\mbox{BS}}_1=0.105$ (SD=0.002), $\widehat{\mbox{BS}}_2=0.042$ (SD=0.001), and $\widehat{\mbox{BS}}_3=0.010$ (SD=0.001), with a global BS of $\widehat{\mbox{BS}}=0.085$ (SD=0.001).

We also evaluate the performance of the non-regularized regression based on the two-stage estimation approach 
\begin{CodeInput}
cross_validator_null = TwoStagesCV()
cross_validator_null.cross_validate(full_df=patients_df, n_splits=4)
\end{CodeInput}
and the global AUCs of the two-stage approach without lasso was $\widehat{\mbox{AUC}}=0.649$ (SD=0.003). Evidently, the global AUCs of the without and with lasso penalty were highly similar.  However, by adding lasso regularization, the number of predictors for each event type is reduced.

\section{Concluding Remarks}
\label{sec:summary}
Discrete-time survival analysis with competing-risks is often required in practical settings. In the friendly package \pkg{PyDTS} we implemented two estimation procedures, that of \cite{lee_analysis_2018} and the faster algorithm of \cite{meir_gorfine_dtsp_2023} which breaks the optimization procedure into a sequence of smaller problems. The faster algorithm could be highly useful in large modern datasets with tens of thousands observations and thousands covariates. \pkg{PyDTS} also provides useful predictions for new set of observations and useful tools for CV model evaluation, grid-search for parameters tuning, and plots. We expect \pkg{PyDTS} to be adopted by \proglang{Python} users and to be further improved by comments and contributions from the  \proglang{Python} community.

\pkg{PyDTS} is an open source \proglang{Python} package which implements tools for discrete-time survival analysis with competing risks. The code is available at https://github.com/tomer1812/pydts \citep{tomer_rom_malka_pydts_2022_git} under GNU GPLv3. Documentation, with details about main functionalities, API, installation and usage examples, is available at \\ https://tomer1812.github.io/pydts/ \citep{tomer_rom_malka_pydts_2022_docs}. The package was developed following best practices of \proglang{Python} programming, automated testing was added for stability, Github Issues can be opened by any user for maintainability and open source community contributions are available for ongoing improvement. 

The package is published using the \proglang{Python} Package Index (PyPI), can be installed (using the “pip” installer) and used with a few simple lines of code, as shown in the Quick Start section of the documentation. Yet, it is flexible enough to support more advanced pre-processing and fitting options. Based on available \proglang{Python} packages (\pkg{Numpy} of \cite{harris2020array}, \pkg{Pandas} of \cite{reback2020pandas}, \pkg{Scipy} of \cite{2020SciPy-NMeth}, \pkg{Scikit-survival} of \cite{sksurv}, \pkg{Lifelines}  of \cite{davidson-pilon_lifelines_2019}, and \pkg{Statsmodels}  of \cite{seabold2010statsmodels}), \pkg{PyDTS} package implements \texttt{DataExpansionFitter} which is the estimation procedure of \cite{lee_analysis_2018} and \texttt{TwoStagesFitter} which is the estimation procedure of \cite{meir_gorfine_dtsp_2023}.

\section*{Data and Code Availability Statement}
Simulated data and code under the GNU GPLv3 are available at the package repository https://github.com/tomer1812/pydts

The MIMIC dataset is accessible at https://physionet.org/content/mimiciv/2.0/ and subjected to PhysioNet credentials.

\section*{Acknowledgements}
The authors would like to thank Hagai Rossman and Ayya Keshet for their helpful discussions and suggestions. M.G. work was supported by the ISF 767/21 grant and Malag competitive grant in data science (DS).

\clearpage

\begin{table}
	\centering
	\caption{Original and expanded datasets with $M=2$ competing events (\cite{lee_analysis_2018})}\label{tbl:expanded}
	\scalebox{1}{
		\begin{tabular}{cccc|cccccc}
			\hline
	\multicolumn{4}{c}{Original Data} & \multicolumn{6}{c}{Expanded Data}  \\
			\midrule
	$i$ & $X_i$ & $\delta_i$ & $Z_i$ & $i$ & $\tilde{X}_i$ & $\delta_{1it}$ &  $\delta_{2it}$ & 
 $1-\delta_{1it}-\delta_{2it}$ & $Z_i$ \\
			\hline
   1 & 2 & 1 & $Z_1$ & 1 & 1 & 0 & 0 & 1 &  $Z_1$ \\
     &   &   &       & 1 & 2 & 1 & 0 & 0 &  $Z_1$ \\
   2 & 3 & 2 & $Z_2$ & 2 & 1 & 0 & 0 & 1 &  $Z_2$ \\
     &   &   &       & 2 & 2 & 0 & 0 & 1 &  $Z_2$ \\    
     &   &   &       & 2 & 3 & 0 & 1 & 0 &  $Z_2$ \\ 
   3 & 3 & 0 & $Z_3$ & 3 & 1 & 0 & 0 & 1 &  $Z_3$ \\
     &   &   &       & 3 & 2 & 0 & 0 & 1 & $Z_3$ \\
     &   &   &       & 3 & 3 & 0 & 0 & 1 & $Z_3$ \\
     \hline
		\end{tabular}
	}
\end{table}

\begin{table}
\centering
\caption{First five rows of a dataframe}
\label{tab:first_observations}
\begin{tabular}{lrrrrrrr}
\toprule
{ID} &  Z1 &     Z2 &  Z3 &  T &  C &  X &  J \\
\midrule
0 &   1 &  97.68 &   4 &  1 &  8 &  1 &  2 \\
1 &   1 &  67.28 &  10 &  8 &  8 &  8 &  0 \\
2 &   1 &  72.61 &   2 &  3 &  7 &  3 &  2 \\
3 &   1 &  80.94 &   6 &  8 &  8 &  8 &  0 \\
4 &   0 &  63.29 &   5 &  8 &  8 &  8 &  0 \\
\bottomrule
\end{tabular}
\end{table}

\begin{table}
\centering
\caption{\label{tab:beta_comparison} Results of one simulated dataset, $n=50,000$: Estimates of $\beta_{j}$.}
\begin{tabular}{lrrrrr}
\toprule
{} &   \multicolumn{1}{l}{True} & \multicolumn{2}{c}{Lee et al.} & \multicolumn{2}{c}{two-step} \\
$\beta_{jk}$ & {} &  Estimate &     SE & Estimate &     SE \\
\midrule
$\beta_{11}$ &  0.223 &      0.193 &  0.026 &    0.188 &  0.025 \\
$\beta_{12}$ & -1.099 &     -1.131 &  0.026 &   -1.101 &  0.026 \\
$\beta_{13}$ & -1.099 &     -1.124 &  0.026 &   -1.093 &  0.026 \\
$\beta_{14}$ & -0.916 &     -0.899 &  0.026 &   -0.875 &  0.025 \\
$\beta_{15}$ & -0.693 &     -0.672 &  0.026 &   -0.653 &  0.025 \\
$\beta_{21}$ & -0.000 &      0.041 &  0.038 &    0.040 &  0.038 \\
$\beta_{22}$ & -1.099 &     -1.113 &  0.039 &   -1.100 &  0.039 \\
$\beta_{23}$ & -1.386 &     -1.426 &  0.040 &   -1.410 &  0.039 \\
$\beta_{24}$ & -1.099 &     -1.111 &  0.039 &   -1.098 &  0.039 \\
$\beta_{25}$ & -0.693 &     -0.662 &  0.039 &   -0.655 &  0.038 \\
\bottomrule
\end{tabular}
\end{table}

\begin{table}
\centering
\caption{\label{tab:rep_beta_comparison} Simulation results based on 100 repetitions: Estimates of $\beta_{j}$ and their standard errors.}
\begin{tabular}{lrrrrr}
\toprule
{} &   \multicolumn{1}{l}{True} & \multicolumn{2}{c}{Lee et al.} & \multicolumn{2}{c}{two-step} \\
$\beta_{jk}$ & {} &  Estimate &     SE & Estimate &     SE \\
\midrule
$\beta_{11}$ &  0.223 &      0.222 &  0.028 &    0.216 &  0.027 \\
$\beta_{12}$ & -1.099 &     -1.100 &  0.033 &   -1.071 &  0.032 \\
$\beta_{13}$ & -1.099 &     -1.098 &  0.029 &   -1.069 &  0.028 \\
$\beta_{14}$ & -0.916 &     -0.916 &  0.030 &   -0.892 &  0.029 \\
$\beta_{15}$ & -0.693 &     -0.691 &  0.029 &   -0.672 &  0.028 \\
$\beta_{21}$ & -0.000 &      0.002 &  0.044 &    0.002 &  0.044 \\
$\beta_{22}$ & -1.099 &     -1.100 &  0.038 &   -1.087 &  0.038 \\
$\beta_{23}$ & -1.386 &     -1.378 &  0.044 &   -1.363 &  0.044 \\
$\beta_{24}$ & -1.099 &     -1.102 &  0.042 &   -1.089 &  0.042 \\
$\beta_{25}$ & -0.693 &     -0.692 &  0.045 &   -0.684 &  0.044 \\
\bottomrule
\end{tabular}
\end{table}

\begin{table}[h!]
\centering
\caption{MIMIC dataset - LOS analysis: Estimated regression coefficients of event type discharge to home, $J=1$.}
\label{table:los-j1}
\begin{tabular}{llccc}
\toprule
{} & {} &     Lee et al. &        Two-Step & Two-Step \& LASSO \\
{} & {} &  Estimate (SE) &   Estimate (SE) &    Estimate (SE) \\
\midrule
Admissions Number  &         2 &   0.000 (0.024) &   0.003 (0.022) &   {\bf 0.000 (0.000)} \\
{} &        3+ &  -0.032 (0.023) &  -0.027 (0.022) &   {\bf 0.000 (0.000)} \\
Anion Gap            &  Abnormal &  -0.137 (0.032) &  -0.128 (0.030) &   {\bf 0.000 (0.000)} \\
Bicarbonate          &  Abnormal &  -0.208 (0.021) &  -0.194 (0.020) &   -0.119 (0.019) \\
Calcium Total        &  Abnormal &  -0.291 (0.020) &  -0.270 (0.019) &   -0.190 (0.018) \\
Chloride             &  Abnormal &  -0.148 (0.024) &  -0.137 (0.023) &   -0.071 (0.021) \\
Creatinine           &  Abnormal &  -0.103 (0.024) &  -0.098 (0.023) &   -0.072 (0.021) \\
Direct Emergency     &       Yes &  -0.011 (0.026) &  -0.014 (0.024) &   {\bf 0.000 (0.000)} \\
Ethnicity      &     Black &   0.006 (0.046) &   0.009 (0.042) &   {\bf 0.000 (0.000)} \\
{}   &  Hispanic &   0.132 (0.053) &   0.120 (0.048) &   {\bf 0.000 (0.000)} \\
{}      &     Other &  -0.162 (0.051) &  -0.146 (0.047) &   {\bf 0.000 (0.000)} \\
{}      &     White &  -0.031 (0.041) &  -0.026 (0.038) &   {\bf 0.000 (0.000)} \\
Glucose              &  Abnormal &  -0.215 (0.018) &  -0.192 (0.016) &   -0.088 (0.016) \\
Hematocrit           &  Abnormal &  -0.042 (0.032) &  -0.037 (0.029) &   -0.042 (0.029) \\
Hemoglobin           &  Abnormal &  -0.080 (0.033) &  -0.071 (0.030) &   -0.081 (0.030) \\
Insurance    &  Medicare &   0.138 (0.039) &   0.125 (0.036) &   {\bf 0.000 (0.000)} \\
{}      &     Other &   0.219 (0.036) &   0.200 (0.033) &    0.030 (0.016) \\
MCH                  &  Abnormal &  -0.002 (0.023) &  -0.002 (0.022) &   {\bf 0.000 (0.000)} \\
MCHC                 &  Abnormal &  -0.128 (0.019) &  -0.116 (0.018) &   -0.003 (0.017) \\
MCV                  &  Abnormal &  -0.048 (0.026) &  -0.045 (0.024) &   {\bf 0.000 (0.000)} \\
Magnesium            &  Abnormal &  -0.080 (0.030) &  -0.074 (0.028) &   {\bf 0.000 (0.000)} \\
Marital Status     &   Married &   0.224 (0.032) &   0.205 (0.030) &    0.093 (0.016) \\
{}       &    Single &  -0.087 (0.033) &  -0.079 (0.031) &   {\bf 0.000 (0.000)} \\
{}      &   Widowed &   0.026 (0.040) &   0.020 (0.037) &   {\bf 0.000 (0.000)} \\
Night Admission      &       Yes &   0.081 (0.017) &   0.075 (0.016) &   {\bf 0.000 (0.000)} \\
Phosphate            &  Abnormal &  -0.052 (0.019) &  -0.048 (0.018) &   {\bf 0.000 (0.000)} \\
Platelet Count       &  Abnormal &  -0.068 (0.019) &  -0.062 (0.018) &   {\bf 0.000 (0.000)} \\
Potassium            &  Abnormal &  -0.103 (0.032) &  -0.095 (0.030) &   {\bf 0.000 (0.000)} \\
RDW                  &  Abnormal &  -0.327 (0.021) &  -0.308 (0.020) &   -0.271 (0.019) \\
Recent Admission     &       Yes &  -0.262 (0.035) &  -0.247 (0.033) &   -0.001 (0.027) \\
Red Blood Cells      &  Abnormal &  -0.089 (0.027) &  -0.078 (0.024) &   -0.024 (0.025) \\
Sex                  &    Female &  -0.007 (0.018) &  -0.006 (0.016) &   {\bf 0.000 (0.000)} \\
Sodium               &  Abnormal &  -0.312 (0.030) &  -0.297 (0.029) &   -0.142 (0.026) \\
Standardized Age     &           &  -0.260 (0.011) &  -0.234 (0.010) &   -0.162 (0.009) \\
Urea Nitrogen        &  Abnormal &  -0.148 (0.022) &  -0.139 (0.020) &   -0.136 (0.020) \\
White Blood Cells    &  Abnormal &  -0.276 (0.018) &  -0.252 (0.016) &   -0.159 (0.016) \\
\bottomrule
\end{tabular}
\end{table}

\begin{table}[h!]
\centering
\caption{MIMIC dataset - LOS analysis: Estimated regression coefficients of event type discharged to another facility, $J=2$.}
\label{table:los-j2}
\begin{tabular}{llccc}
\toprule
{} &  {} &    Lee et al. &        Two-Step & Two-Step \& LASSO \\
{} &  {} & Estimate (SE) &   Estimate (SE) &    Estimate (SE) \\
\midrule
Admissions Number   &         2 &   0.108 (0.041) &   0.107 (0.040) &    0.087 (0.038) \\
{}  &        3+ &   0.194 (0.037) &   0.190 (0.036) &    0.169 (0.034) \\
Anion Gap            &  Abnormal &  -0.006 (0.048) &  -0.006 (0.047) &    {\bf 0.000 (0.002)} \\
Bicarbonate          &  Abnormal &  -0.121 (0.033) &  -0.117 (0.032) &   -0.110 (0.032) \\
Calcium Total        &  Abnormal &  -0.098 (0.031) &  -0.094 (0.031) &   -0.088 (0.030) \\
Chloride             &  Abnormal &   0.016 (0.036) &   0.015 (0.035) &    {\bf 0.000 (0.002)} \\
Creatinine           &  Abnormal &  -0.199 (0.036) &  -0.191 (0.035) &   -0.173 (0.035) \\
Direct Emergency     &       Yes &  -0.373 (0.052) &  -0.363 (0.050) &   -0.345 (0.050) \\
Ethnicity       &     Black &   0.084 (0.090) &   0.079 (0.088) &    0.028 (0.086) \\
{}   &  Hispanic &  -0.068 (0.111) &  -0.070 (0.108) &   -0.088 (0.106) \\
{}     &     Other &   0.026 (0.099) &   0.022 (0.097) &   -0.006 (0.095) \\
{}      &     White &   0.144 (0.082) &   0.138 (0.081) &    0.094 (0.079) \\
Glucose              &  Abnormal &  -0.138 (0.031) &  -0.132 (0.030) &   -0.126 (0.030) \\
Hematocrit           &  Abnormal &   0.038 (0.057) &   0.039 (0.055) &    0.032 (0.055) \\
Hemoglobin           &  Abnormal &   0.018 (0.062) &   0.015 (0.060) &    0.005 (0.059) \\
Insurance    &  Medicare &   0.237 (0.075) &   0.230 (0.074) &    0.238 (0.073) \\
{}     &     Other &  -0.094 (0.074) &  -0.091 (0.072) &   -0.081 (0.072) \\
MCH                  &  Abnormal &   0.042 (0.038) &   0.040 (0.037) &    0.019 (0.031) \\
MCHC                 &  Abnormal &  -0.010 (0.031) &  -0.011 (0.030) &    {\bf 0.000 (0.003)} \\
MCV                  &  Abnormal &  -0.020 (0.041) &  -0.019 (0.039) &    {\bf 0.000 (0.003)} \\
Magnesium            &  Abnormal &  -0.039 (0.048) &  -0.038 (0.047) &   -0.025 (0.046) \\
Marital Status      &   Married &  -0.254 (0.054) &  -0.249 (0.053) &   -0.262 (0.052) \\
{}      &    Single &   0.209 (0.054) &   0.200 (0.053) &    0.176 (0.052) \\
{}      &   Widowed &   0.175 (0.058) &   0.163 (0.056) &    0.149 (0.056) \\
Night Admission      &       Yes &   0.056 (0.029) &   0.054 (0.028) &    0.047 (0.028) \\
Phosphate            &  Abnormal &  -0.042 (0.033) &  -0.040 (0.032) &   -0.034 (0.031) \\
Platelet Count       &  Abnormal &  -0.130 (0.032) &  -0.125 (0.031) &   -0.118 (0.031) \\
Potassium            &  Abnormal &   0.042 (0.048) &   0.042 (0.047) &    0.023 (0.047) \\
RDW                  &  Abnormal &  -0.107 (0.033) &  -0.104 (0.032) &   -0.093 (0.031) \\
Recent Admission     &       Yes &  -0.021 (0.051) &  -0.023 (0.049) &    {\bf 0.000 (0.004)} \\
Red Blood Cells      &  Abnormal &   0.083 (0.052) &   0.079 (0.050) &    0.073 (0.050) \\
Sex                  &    Female &   0.090 (0.031) &   0.088 (0.030) &    0.078 (0.030) \\
Sodium               &  Abnormal &  -0.056 (0.042) &  -0.056 (0.041) &   -0.039 (0.038) \\
Standardized Age     &           &   0.536 (0.021) &   0.525 (0.021) &    0.519 (0.021) \\
Urea Nitrogen        &  Abnormal &   0.100 (0.035) &   0.095 (0.034) &    0.077 (0.034) \\
White Blood Cells    &  Abnormal &  -0.107 (0.029) &  -0.103 (0.028) &   -0.099 (0.028) \\
\bottomrule
\end{tabular}
\end{table}

\begin{table}[h!]
\centering
\caption{MIMIC dataset - LOS analysis: Estimated regression coefficients of event type in-hospital death, $J=3$.}
\label{table:los-j3}
\begin{tabular}{llccc}
\toprule
{} & {} &     Lee et al. &        Two-Step & Two-Step \& LASSO \\
{} & {} &  Estimate (SE) &   Estimate (SE) &    Estimate (SE) \\
\midrule
Admissions Number   &         2 &   0.147 (0.074) &   0.147 (0.073) &    0.140 (0.074) \\
{}  &        3+ &   0.142 (0.069) &   0.140 (0.068) &    0.134 (0.068) \\
Anion Gap            &  Abnormal &   0.582 (0.064) &   0.573 (0.064) &    0.571 (0.064) \\
Bicarbonate          &  Abnormal &   0.543 (0.056) &   0.537 (0.056) &    0.535 (0.056) \\
Calcium Total        &  Abnormal &   0.204 (0.054) &   0.204 (0.054) &    0.203 (0.054) \\
Chloride             &  Abnormal &   0.147 (0.059) &   0.143 (0.058) &    0.142 (0.058) \\
Creatinine           &  Abnormal &   0.273 (0.067) &   0.271 (0.067) &    0.271 (0.067) \\
Direct Emergency     &       Yes &  -0.318 (0.096) &  -0.311 (0.095) &   -0.302 (0.095) \\
Ethnicity       &     Black &  -0.236 (0.140) &  -0.235 (0.139) &   -0.203 (0.140) \\
{}    &  Hispanic &  -0.395 (0.183) &  -0.393 (0.181) &   -0.351 (0.181) \\
{}      &     Other &   0.145 (0.147) &   0.133 (0.145) &    0.155 (0.146) \\
{}      &     White &  -0.156 (0.123) &  -0.157 (0.122) &   -0.130 (0.123) \\
Glucose              &  Abnormal &   0.215 (0.064) &   0.212 (0.063) &    0.208 (0.063) \\
Hematocrit           &  Abnormal &  -0.198 (0.108) &  -0.194 (0.107) &   -0.165 (0.108) \\
Hemoglobin           &  Abnormal &   0.024 (0.122) &   0.023 (0.121) &    0.003 (0.121) \\
Insurance    &  Medicare &  -0.224 (0.136) &  -0.225 (0.135) &   -0.171 (0.138) \\
{}      &     Other &  -0.242 (0.133) &  -0.240 (0.132) &   -0.188 (0.135) \\
MCH                  &  Abnormal &  -0.066 (0.070) &  -0.066 (0.069) &   -0.057 (0.069) \\
MCHC                 &  Abnormal &   0.027 (0.056) &   0.029 (0.055) &    0.027 (0.055) \\
MCV                  &  Abnormal &   0.060 (0.072) &   0.061 (0.071) &    0.055 (0.071) \\
Magnesium            &  Abnormal &   0.329 (0.073) &   0.324 (0.072) &    0.320 (0.072) \\
Marital Status      &   Married &   0.156 (0.102) &   0.154 (0.101) &    0.127 (0.061) \\
{}       &    Single &   0.026 (0.107) &   0.027 (0.106) &   {\bf 0.000 (0.008)} \\
{}     &   Widowed &   0.047 (0.115) &   0.048 (0.114) &    0.020 (0.084) \\
Night Admission      &       Yes &  -0.096 (0.053) &  -0.093 (0.052) &   -0.089 (0.052) \\
Phosphate            &  Abnormal &   0.178 (0.056) &   0.176 (0.055) &    0.174 (0.055) \\
Platelet Count       &  Abnormal &   0.235 (0.054) &   0.232 (0.054) &    0.229 (0.054) \\
Potassium            &  Abnormal &   0.227 (0.072) &   0.221 (0.071) &    0.221 (0.071) \\
RDW                  &  Abnormal &   0.492 (0.058) &   0.486 (0.058) &    0.483 (0.058) \\
Recent Admission     &       Yes &   0.250 (0.083) &   0.242 (0.082) &    0.242 (0.082) \\
Red Blood Cells      &  Abnormal &   0.142 (0.105) &   0.140 (0.104) &    0.130 (0.104) \\
Sex                  &    Female &  -0.011 (0.057) &  -0.008 (0.057) &   -0.005 (0.057) \\
Sodium               &  Abnormal &   0.276 (0.064) &   0.270 (0.063) &    0.268 (0.063) \\
Standardized Age     &           &   0.580 (0.041) &   0.574 (0.040) &    0.568 (0.040) \\
Urea Nitrogen        &  Abnormal &   0.141 (0.070) &   0.141 (0.070) &    0.141 (0.070) \\
White Blood Cells    &  Abnormal &   0.579 (0.056) &   0.571 (0.056) &    0.568 (0.055) \\
\bottomrule
\end{tabular}
\end{table}

\clearpage

\clearpage
\bibliographystyle{plainnat}
\bibliography{pydts}

@software{tomer_rom_malka_pydts_2022_git,
    author = {Meir, Tomer and Gutman, Rom and Gorfine, Malka},
    %doi = {},
    license = {MIT},
    month = {4},
    title = {{PyDTS - Python Package for Discrete Time Survival-analysis - Github Repository}},
    url = {https://github.com/tomer1812/pydts},
    year = {2022}
}

@software{tomer_rom_malka_pydts_2022_docs,
    author = {Meir, Tomer and Gutman, Rom and Gorfine, Malka},
    %doi = {},
    license = {MIT},
    month = {4},
    title = {{PyDTS - Python Package for Discrete Time Survival-analysis - Documentation}},
    url = {https://tomer1812.github.io/pydts/},
    year = {2022}
}

@article{sksurv,
  author  = {Sebastian P{\"o}lsterl},
  title   = {scikit-survival: A Library for Time-to-Event Analysis Built on Top of scikit-learn},
  journal = {Journal of Machine Learning Research},
  year    = {2020},
  volume  = {21},
  number  = {212},
  pages   = {1-6},
  url     = {http://jmlr.org/papers/v21/20-729.html}
}

@software{reback2020pandas,
    author       = {Jeff Reback; jbrockmendel; Wes McKinney; Joris Van den Bossche; Tom Augspurger; Phillip Cloud; Simon Hawkins; Matthew Roeschke; gfyoung; Sinhrks; Adam Klein; Patrick Hoefler; Terji Petersen; Jeff Tratner; Chang She; William Ayd; Shahar Naveh; JHM Darbyshire; Marc Garcia; Richard Shadrach; Jeremy Schendel; Andy Hayden; Daniel Saxton; Marco Edward Gorelli; Fangchen Li; Matthew Zeitlin; Vytautas Jancauskas; Ali McMaster; Pietro Battiston; Skipper Seabold},
    title        = {pandas-dev/pandas: Pandas},
    month        = feb,
    year         = 2020,
    publisher    = {Zenodo},
    version      = {latest},
    doi          = {10.5281/zenodo.6053272},
}

@Article{         harris2020array,
 title         = {Array programming with {NumPy}},
 author        = {Charles R. Harris and K. Jarrod Millman and St{\'{e}}fan J.
                 van der Walt and Ralf Gommers and Pauli Virtanen and David
                 Cournapeau and Eric Wieser and Julian Taylor and Sebastian
                 Berg and Nathaniel J. Smith and Robert Kern and Matti Picus
                 and Stephan Hoyer and Marten H. van Kerkwijk and Matthew
                 Brett and Allan Haldane and Jaime Fern{\'{a}}ndez del
                 R{\'{i}}o and Mark Wiebe and Pearu Peterson and Pierre
                 G{\'{e}}rard-Marchant and Kevin Sheppard and Tyler Reddy and
                 Warren Weckesser and Hameer Abbasi and Christoph Gohlke and
                 Travis E. Oliphant},
 year          = {2020},
 month         = sep,
 journal       = {Nature},
 volume        = {585},
 number        = {7825},
 pages         = {357--362},
 doi           = {10.1038/s41586-020-2649-2},
 publisher     = {Springer Science and Business Media {LLC}},
}

@inproceedings{seabold2010statsmodels,
  title={statsmodels: Econometric and statistical modeling with python},
  author={Seabold, Skipper and Perktold, Josef},
  booktitle={9th Python in Science Conference},
  year={2010},
}

@ARTICLE{2020SciPy-NMeth,
  author  = {Virtanen, Pauli and Gommers, Ralf and Oliphant, Travis E. and
            Haberland, Matt and Reddy, Tyler and Cournapeau, David and
            Burovski, Evgeni and Peterson, Pearu and Weckesser, Warren and
            Bright, Jonathan and {van der Walt}, St{\'e}fan J. and
            Brett, Matthew and Wilson, Joshua and Millman, K. Jarrod and
            Mayorov, Nikolay and Nelson, Andrew R. J. and Jones, Eric and
            Kern, Robert and Larson, Eric and Carey, C J and
            Polat, {\.I}lhan and Feng, Yu and Moore, Eric W. and
            {VanderPlas}, Jake and Laxalde, Denis and Perktold, Josef and
            Cimrman, Robert and Henriksen, Ian and Quintero, E. A. and
            Harris, Charles R. and Archibald, Anne M. and
            Ribeiro, Ant{\^o}nio H. and Pedregosa, Fabian and
            {van Mulbregt}, Paul and {SciPy 1.0 Contributors}},
  title   = {{{SciPy} 1.0: Fundamental Algorithms for Scientific
            Computing in Python}},
  journal = {Nature Methods},
  year    = {2020},
  volume  = {17},
  pages   = {261--272},
  adsurl  = {https://rdcu.be/b08Wh},
  doi     = {10.1038/s41592-019-0686-2},
}

@article{tibshirani1996regression,
  title={Regression Shrinkage and Selection via the Lasso},
  author={Tibshirani, Robert},
  journal={Journal of the Royal Statistical Society: Series B (Methodological)},
  volume={58},
  number={1},
  pages={267--288},
  year={1996},
  url={https://www.jstor.org/stable/2346178}
}

@article{hoel1970ridge,
  title={Ridge Regression: Biased Estimation for Nonorthogonal Problem.},
  author={Hoel, Arthur E and Kennard, Robert W},
  journal={Technometrics},
  volume={42},
  number={1},
  pages={80--86},
  year={1970},
  doi={10.2307/1271436}
}

@article{zou2005regularization,
  title={Regularization and Variable Selection via the Elastic Net},
  author={Zou, Hui and Hastie, Trevor},
  journal={Journal of the Royal Statistical Society: Series B (Statistical Methodology)},
  volume={67},
  number={2},
  pages={301--320},
  year={2005},
  url = {https://www.jstor.org/stable/3647580}
}

@book{therneau2000cox,
  title={Modeling Survival Data: Extending the Cox Model.},
  author={Therneau, Terry M and Grambsch, Patricia M},
  year={2000},
  publisher={Springer-Verlag},
  doi={10.1007/978-1-4757-3294-8}
}

@article{wu2022analysis,
  title={Analysis of Hospital Readmissions with Competing Risks},
  author={Wu, Wenbo and He, Kevin and Shi, Xu and Schaubel, Douglas E and Kalbfleisch, John D},
  journal={Statistical Methods in Medical Research},
  pages={2189--2200},
  volume={31},
  number={11},
  year={2022},
  doi={10.1177/09622802221115879},
}

@Manual{discSurv22,
    title = { {discSurv}: Discrete Time Survival Analysis},
    author = {Thomas Welchowski and Moritz Berger and David Koehler and Matthias Schmid},
    year = {2022},
    note = { {R}~package version~2.0.0},
    url = {https://CRAN.R-project.org/package=discSurv},
  }

@book{hastie2009elements,
  title={The Elements of Statistical Learning: Data Mining, Inference, and Prediction.},
  author={Hastie, Trevor and Tibshirani, Robert and Friedman, Jerome H},
  Edition={2nd},
  year={2009},
  publisher={Springer-Verlag},
  doi={10.1007/978-0-387-84858-7}
}

@article{berger2019classification,
  title={A Classification Tree Approach for the Modeling of Competing Risks in Discrete Time},
  author={Berger, Moritz and Welchowski, Thomas and Schmitz-Valckenberg, Steffen and Schmid, Matthias},
  journal={The Advances in Data Analysis and Classification},
  volume={13},
  number={4},
  pages={965--990},
  year={2019},
  publisher={Springer-Verlag},
  doi = {10.1007/s11634-018-0345-y}
}

@article{schmid2021competing,
  title={Competing Risks Analysis for Discrete Time-to-Event Data},
  author={Schmid, Matthias and Berger, Moritz},
  journal={WIREs Computational Statistics},
  volume={13},
  number={5},
  pages={e1529},
  year={2021},
  doi={10.1002/wics.1529},
}

@book{klein_survival_2003,
	title = {Survival Analysis},
	publisher = {Springer-Verlag},
	author = {Klein, John P. and Moeschberger, Melvin L.},
	date = {2003},
	year = {2003},
        doi = {10.1007/b97377}
 }

@book{kalbfleisch_statistical_2011,
	edition = {2nd},
	title = {The Statistical Analysis of Failure Time Data},
	publisher = {John Wiley \& Sons},
	author = {Kalbfleisch, John D. and Prentice, Ross L.},
	date = {2011-01},
	year = {2011},
        doi = {10.1002/9781118032985}
}

@article{cox_regression_1972,
	title = {Regression Models and Life-Tables.},
	volume = {34},
	doi = {10.1111/j.2517-6161.1972.tb00899.x},
	pages = {187--202},
	number = {2},
	journaltitle = {Journal of the Royal Statistical Society: Series B (Methodological)},
	shortjournal = {Journal of the Royal Statistical Society: Series B (Methodological)},
	author = {Cox, D. R.},
	date = {1972-01},
	year = {1972},
	journal = {Journal of the Royal Statistical Society: Series B (Methodological)}
}

@article{efron_efficiency_1977,
	title = {The Efficiency of Cox's Likelihood Function for Censored Data},
	volume = {72},
	doi = {10.1080/01621459.1977.10480613},
	pages = {557--565},
	number = {359},
	journaltitle = {Journal of the American Statistical Association},
	shortjournal = {Journal of the American Statistical Association},
	author = {Efron, Bradley},
	date = {1977-09},
	year = {1977}
}

@article{lee_analysis_2018,
	title = {On the Analysis of Discrete Time Competing Risks Data},
	volume = {74},
	doi = {10.1111/biom.12881},
	pages = {1468--1481},
	number = {4},
	journaltitle = {Biometrics},
	shortjournal = {Biom},
	author = {Lee, Minjung and Feuer, Eric J. and Fine, Jason P.},
	date = {2018-12},
	year = {2018},
	journal = {Biometrics}
}

@article{allison_discrete-time_1982,
    title = {Discrete-Time Methods for the Analysis of Event Histories.},
    volume = {13},
    number = {},
    doi = {10.2307/270718},
    pages = {61--98},
    journaltitle = {Sociological Methodology},
    shortjournal = {Sociological Methodology},
    author = {Allison, Paul D.},
    date = {1982},
    year = {1982},
    journal = {Sociological Methodology},
}

@article{davidson-pilon_lifelines_2019,
	title = { {lifelines}: Survival Analysis in  {Python}},
	volume = {4},
	doi = {10.21105/joss.01317},
	shorttitle = {lifelines},
	pages = {1317},
	number = {40},
	journaltitle = {Journal of Open Source Software},
	shortjournal = {{JOSS}},
	author = {Davidson-Pilon, Cameron},
	date = {2019-08-04},
	year = {2019},
	journal = {Journal of Open Source Software}
}

@article{johnson_mimic-iv_2022,
	title = {{MIMIC}-{IV} (version 2.0)},
	doi = {10.13026/7vcr-e114},
	journal = {PhysioNet},
	author = {Johnson, Alistair and Bulgarelli, Lucas and Pollard, Tom and Horng, Steven and Celi, Leo Anthony and Mark, Roger},
	month = jun,
	year = {2022},
}

@article{goldberger_physiobank_2000,
	title = {PhysioBank, PhysioToolkit, and PhysioNet: Components of a New Research Resource for Complex Physiologic Signals},
	volume = {101},
	shorttitle = {PhysioBank, PhysioToolkit, and PhysioNet},
	doi = {10.1161/01.CIR.101.23.e215},
	number = {23},
	journal = {Circulation},
	author = {Goldberger, Ary L. and Amaral, Luis A. N. and Glass, Leon and Hausdorff, Jeffrey M. and Ivanov, Plamen Ch. and Mark, Roger G. and Mietus, Joseph E. and Moody, George B. and Peng, Chung-Kang and Stanley, H. Eugene},
	year = {2000},
}

@article{meir_gorfine_dtsp_2023,
    author = {Meir, Tomer and Gorfine, Malka},
    doi = {10.1093/biomtc/ujaf040},
    title = {{Discrete-Time Competing-Risks Regression with or without Penalization}},
    year = {2025},
    journal = {Biometrics},
    volume = {81},
    issue = {2}
}
\end{document}